# Translating Regulatory Clauses into Executable Codes for Building Design Checking via Large Language Model Driven Function Matching and Composing

Zhe Zheng[1, 2,§], Jin Han[1,§], Ke-Yin Chen[1], Xin-Yu Cao[3], Xin-Zheng Lu[1], Jia-Rui Lin[1,4,*]

1. Department of Civil Engineering, Tsinghua University, Beijing, 100084, China

2. Dept. of Investment and Technology Innovation, Wuliangye Yibin Co., Ltd.

3. School of Civil Engineering, Yantai University, Yantai 264005, China

4. Key Laboratory of Digital Construction and Digital Twin, Ministry of Housing and Urban-Rural Development, China

[§]These authors contributed equally to this manuscript.

*Corresponding author, E-mail: lin611@tsinghua.edu.cn

**Abstract:**

Translating clauses into executable code is a vital stage of automated rule checking (ARC) and is essential for effective building design compliance checking, particularly for rules with implicit properties or complex logic requiring domain knowledge. Thus, by systematically analyzing building clauses, 66 atomic functions are defined first to encapsulate common computational logics. Then, LLM-FuncMapper is proposed, a large language model (LLM)-based approach with rule-based adaptive prompts that match clauses to atomic functions. Finally, executable code is generated by composing functions through the LLMs. Experiments show LLM-FuncMapper outperforms fine-tuning methods by 19% in function matching while significantly reducing manual annotation efforts. A case study demonstrates that LLM-FuncMapper can automatically compose multiple atomic functions to generate executable code, boosting rule-checking efficiency. To our knowledge, this research represents the first application of LLMs for interpreting complex design clauses into executable code, which may shed light on further adoption of LLMs in the construction domain.

# 1 Introduction

Buildings in the entire lifecycle, including design, construction, operation, and maintenance stages, must comply with the requirements of the various building codes to ensure safety, sustainability, and comfort (Eastman et al., 2009; Soliman-Junior et al.,2021). However, the manual checking process is inadequate for dealing with the massive and complex building codes, as it is time-consuming, costly, and error-prone (Zhang and El-Gohary, 2017; Xu and Cai, 2021). To enhance the efficiency and reliability of rule checking, automated rule checking (ARC) has been widely studied in the architectural, engineering, and construction (AEC) field over the past decades (Zhong et al., 2020). Most of the existing ARC systems still require extensive manual work for rule interpretation and rule input (Nawari, 2020). For example, both the first rule checking system CORENET, and the widely adopted Solibri Model Checker (SMC) (Solibri Model Checker, 2023) employ hard-coded rules and rely on manual interpretation of regulatory clauses (Eastman et al., 2009). However, this method entails embedding the regulatory clauses directly in the rule engine, and any changes in the design rules require manual modifications by the domain experts, which limits its adaptability to different domains (Dimyadi et al., 2016; Nawari, 2019, 2020). Therefore, the rule interpretation and code generation stage is now widely recognized as the most critical and challenging step toward fully automated rule checking (Ismail et al., 2017).

Semi-automated rule interpretation methods and Natural Language Processing (NLP)-based automated rule interpretation methods have been studied recently (Fuchs & Amor, 2021; Zheng et al., 2022a; Zhou et al., 2022). Although these studies have advanced the development of ARC, these methods can only handle text at a coarse-grained level (i.e., providing a single label to a sentence or a long phrase instead of each word or concept in a sentence) and require manually labeling a large number of regulatory documents and generating pseudocodes from them (Nawari, 2019). Therefore, NLP techniques are employed by researchers to explore automated rule interpretation methods. However, most of the existing automated rule interpretation methods mainly focused on checking simple clauses which only involved the explicitly stored properties (e.g., attributes and entity references), which fall into Class1 (i.e., the rules that require a single or small number of explicit data) according to the rule complexity defined by Solihin and Eastman (Solihin and Eastman, 2015). This is mainly because interpreting complex clauses requires the implicit properties to be calculated from the existing properties via complex computational logics, including mathematical and geometric operations. However, first-order logics that are often used for automated rule interpretation (Ismail et al., 2017), such as the Horn clause (Zhang and El-Gohary, 2016), B-Prolog representation (Zhang and El-Gohary, 2015; Zhou and El-Gohary, 2017), or the deontic logic (DL) clauses (Xu and Cai, 2021), are still limited in expressibility (Kuske and Schweikardt, 2017). These logic representation methods struggle to describe the clauses with implicit properties that demand complex computation logics (e.g., counting quantifiers, computational geometry) (Kuske and Schweikardt, 2017; Zheng et al., 2022a).

Two types of approaches, domain-specific query language (DSQL) and high-level function library, respectively, have been introduced to represent the complex computational logic within clauses using domain-specific calculation functions. However, function library is in their nascent stages and have only been explored in dealing with Korean building codes. Further research is still needed to explore the adaptability of similar approaches to building codes in other countries to facilitate automated rule interpretation.

Additionally, within the above methods, rule interpretation still highly depends on extensive manual efforts, which is time-consuming and far from automated rule interpretation. Only the professionals possessing extensive expertise are able to choose the proper function from a vast array of predefined DSQLs or function library according to the semantics of clauses. Therefore, computer-aided methods to reduce the difficulty of function selection are needed. Nevertheless, choosing the proper function requires a comprehensive understanding of the semantics of clauses, which is hard for computers. Recently, LLMs have shown a more promising ability in natural language understanding than other methods (Han et al., 2025). While these studies achieve high precision in translating regulations, the interpretation of complex clauses that involve multiple levels or require additional computation remains an area for further investigation. Overall, how to integrate domain-specific knowledge into LLMs to achieve function matching from complex clauses and compose them into executable code, thereby facilitating rule interpretation, remains an open research question.

Therefore, to address the above problems, this work presents LLM-FuncMapper, a method to match clauses to atomic functions and then compose them into executable code based on the large language model (LLM). The method consists of the establishment of atomic functions to capture shared computational logic and a rule-based adaptive prompt engineering to enable LLMs in function matching and code generation.

The remainder of this paper is organized as follows. Section 2 reviews the related studies and highlights the potential research gaps. Section 3 illustrates the methodology of LLM-FuncMapper. Section 4 conducts descriptive and statistical analyses of the proposed atomic function library. Section 5 performs experiments to show the performance of the proposed LLM-FuncMapper in atomic function matching. Section 6 conducts a rule checking of an actual plant as a proof of concept to validate the effectiveness of the proposed method in interpreting complex design clauses into executable code through the composition of atomic functions. Section 7 highlights the insights of this research and also summarizes potential limitations. Finally, Section 8 concludes this research.

# 2 Related work

## 2.1 Automated rule interpretation

The rule interpretation stage is considered to be the most important and complex stage in the process of achieving fully automated rule checking (Ismail et al., 2017). A number of researchers have explored semi-automated rule interpretation methods and automated rule interpretation methods.

The semi-automated rule interpretation method aims to formalize the clauses and thus simplify the rule interpretation process. An eXtensible Markup Language (XML)-based document representation method is proposed to help user understanding and computational analysis (Lau and Law, 2004). The RASE method (Hjelseth and Nisbet, 2011) can process building codes with markup based on the four operators to generate testable logical statements on different types of regulatory clauses. However, the markup process is a manual process. Beach et al. (Beach et al., 2015) extended the RASE method by mapping the documents labeled by the RASE method to the Semantic Web Rule Language (SWRL).

Solihin and Eastman (Solihin and Eastman, 2016) introduced the conceptual graph (CG) to represent the building codes to facilitate rule interpretation. Although these methods achieve semi-automated rule interpretation and improve the accuracy of rule interpretation, they can only process text at a coarse-grained level, which will lead to difficulty in semantic alignment with elements in BIM models. Besides, labeling regulatory documents will involve a lot of manual work.

The NLP-based automated rule interpretation method aims to completely eliminate the reliance on manual work, using NLP techniques to enable computers to handle the semantics of natural language from building codes and then complete rule interpretation (Song et al., 2018). Zhang and EI-Gohary proposed methods to capture syntactic features and semantic features using rule-based NLP techniques and domain ontology to support the automated extraction of information from regulatory documents (Zhang and El-Gohary, 2015; Zhang and El-Gohary, 2016). de Moreira Bohnenberger utilizes semantic similarity measures and natural language processing to concurrently extract process information from multiple documents (de Moreira Bohnenberger et al., 2024). Word embedding associated with deep learning models is utilized to filter irrelevant sentences in the Korean building code to support automated rule interpretation (Song et al., 2018). LSTM-based methods are proposed to automatically extract semantic elements (Moon et al., 2022) and semantic relations between the elements (Zhang and El-Gohary, 2022) to achieve rule interpretation. Zhou et al. used deep learning models (BERT) for semantic annotation and proposed syntactic parsing methods that can automatically convert the input token into a rule check tree (RCTree) (Zhou et al., 2022). Based on this, Zheng et al. proposed an unsupervised learning-based semantic alignment method and a knowledge-based conflict resolution to improve the accuracy of rule interpretation (Zheng et al., 2022a). Additionally, scholars have also attempted to translate natural language legal requirements from various domains into formal representations (Fuchs et al., 2024a; Fuchs et al., 2024b; Manas et al, 2024; Zhang, 2023). However, most of the existing automated rule interpretation methods lack the incorporation of domain knowledge or mainly focus on simple clauses and seldom focus on interpreting the complex clauses that require complex computational mathematical or geometric logic. One of the issues is that the widely-used first-order logic struggles to describe the clauses with implicit properties that demand complex computation logic (Kuske and Schweikardt, 2017; Zheng et al., 2022a). For instance, consider the clause: "Adjacent nursing units in hospitals shall be separated by fire partition walls with a fire resistance rating of not less than 2.00h (from Chinese building code)", because the topology relationships (i.e., adjacent) are not explicitly stored in models and should be derived from geometry information, it would be challenging to check the adjacency relationship between nursing units and to accurately locate the position of the partition wall using first-order logic. For another clause: "The distance from any point in the plant to the safety exit should not be larger than 50m", because the geometry information (i.e., distance from any point to the safety exit) is not explicitly stored, it would be challenging to check only using first-order logic. Therefore, domain-specific calculation functions should be introduced to interpret complex rules with complex computational logic.

## 2.2 Computational logic representation

First-order logic and SPARQL are widely used for the rule interpreting process. However, their expressibility is limited; thus, it is difficult to describe the clauses with complex computation logics.

Therefore, domain-specific calculation functions should be introduced to interpret complex rules with complex computational logic. The research on the domain-specific calculation functions includes two main types, which are DSQL and high-level function library, as summarized in Table 1.

To reduce the difficulty of querying building information, DSQLs have been extensively studied (Mazairac and Beetz, 2013; Daum and Borrmann, 2013, 2014, 2015). DSQL simplifies queries by introducing specific query keywords and corresponding specific functions (operators). For example, Mazairac and Beetz proposed BIMQL, containing query keywords including 'select' and 'where', etc., for selecting and updating data stored in Industry Foundation Classes models (Mazairac and Beetz, 2013). However, BIMQL does not support spatial queries. Daum and Borrmann proposed that the QL4BIM language provides metric, directional, and topological operators to express clauses with qualitative spatial semantics (Daum and Borrmann, 2013, 2014, 2015). Zhang et al. proposed BimSPARQL, a method for extending domain-specific functions in SPARQL to apply spatial and logical reasoning to simplify writing queries and enhance query abilities (Zhang et al., 2018). In total, the BimSPARQL method introduces 1896 functions, which may pose a heavy burden for engineers to find proper functions. Sydora and Stroulia proposed a domain-specific language for computationally representing building interior design rules (Sydora and Stroulia, 2020).

In addition to DSQL, very few studies focused on building library of high-level functions to represent the complex computational logic embedded in clauses. Typical examples are the function library based on requests for proposals for building designs in South Korea (Uhm et al., 2015) and the function library based on the Korean building act (Lee et al., 2023). The functions in the library support complex computational logic such as topological checking, complex geometric checking, complex entity relationship checking, and so on.

Despite the existing efforts, some DSQLs only focused on partial computational logic (e.g., not supporting topology or geometry querying), which restricted their application. The function library usually contain comprehensive functions for most kinds of computational logic within the target codes. However, the existing studies on function library mainly focus on Korean building codes. The applicability and expressibility of the aforementioned function library in building codes in other countries still need to be studied. Although some complex computational logic embedded in clauses can be expressed using DSQL and function library, these methods have high learning costs (Mernik et al., 2005; Zhou et al., 2022) and require human efforts for searching proper functions (e.g., finding one proper function from 1896 functions defined by BimSPARQL).

Table 1 Computational logic representation methods

| Reference | Representation methods | Country |
|---|---|---|
| Mazairac and Beetz, 2013 | BIMQL | Netherlands |
| Zhang et al., 2018 | BimSPARQL | USA |
| Daum and Borrmann, 2015 | QL4BIM | Germany |
| Sydora and Stroulia, 2020 | Domain language for interior design | Canada |
| Uhm et al., 2015 | High-level function library | Korea |
| Lee et al., 2023 | High-level function library | Korea |

## 2.3 Function matching

Each function is a highly encapsulated method for data retrieval, reasoning, or computation provided by predefined function library. Domain experts can use these functions to reduce the difficulty of rule interpretation. However, in the rule interpreting stage, domain experts may spend lots of time selecting the most appropriate functions for each clause from the vast number of functions provided by the function library. Thus, the automated function matching method for clauses is urgently needed. However, few studies focused on function matching methods in the AEC domain. The function matching task is similar to the Application Programming Interface (API) task recommendation in the computer science domain. A number of studies have been devoted to improving the automated recognition of APIs (Peng et al., 2023), and the recommendation of APIs has gone through the following stages.

Initially, models based on probability and statistics (e.g., N-gram) or data mining methods (e.g., frequent pattern mining) were used to learn API usage patterns from large-scale codebases. And then these models and patterns can be used to recommend APIs. However, these approaches cannot take semantic information into account, and cannot handle multiple or cross-database cases (Nguyen et al., 2016; Zhong et al., 2009). With the development of deep learning, some research efforts have been devoted to using deep neural networks (e.g., RNN, LSTM, Transformer, BERT, etc.) to model API sequences. These methods can improve the accuracy and generalization of API recommendations with superior performance (Peng et al., 2023). In the last two years, due to the worldwide popularity of LLMs, the researchers have started to explore the adoption (e.g., ChatGPT (OpenAI, 2022), LLaMA (Touvron et al., 2023)) for API recommendation. For example, Patil et al. (Patil et al., 2023) proposed a finetuned LLaMA-based model, Gorilla, which surpasses the performance of GPT-4 in API calls in massive dataset tests and can also support real-time updates of documents, improving the accuracy and reliability of API calls. Besides, some research on LLMs within the ARC has focused on translating natural language legal requirements into formal representations. For instance, Fuchs (Fuchs et al., 2024a; Fuchs et al., 2024b) utilized LLMs to convert building regulations into formal representations using LegalRuleML. Similarly, Manas (Manas et al, 2024) applied LLMs to automatically translate traffic rules into metric temporal logic (MTL). Furthermore, Zhang (Zhang, 2023) demonstrated the potential of ChatGPT to convert building code requirements into Python code. However, as demonstrated by Zhang (Zhang, 2023), this step faces challenges due to the implicit semantics, domain-specific concepts, and complex logical structures embedded in regulatory texts. Many design rules involve nuanced domain-specific knowledge, conditional constraints, or interdependent clauses that are difficult to formalize. As a result, accurate code generation requires natural language understanding and the ability to match regulatory intent to precise, modular rule logic, making it a key bottleneck in pursuing fully automated and generalizable rule-checking solutions.

## 2.4 Research gaps

Although lots of efforts have been made to represent complex computational logic within clauses for automated rule interpretation, there are still some limitations in the following three main aspects.

First, most of the existing automated rule interpretation methods can only analyze and interpret simple clauses because the widely used first-order logic struggles to describe the clauses with implicit properties that demand complex computation logic. Second, the studies on representing complex computational logic mainly include DSQLs and high-level function library. Within the above methods, rule interpretation still highly depends on extensive manual efforts, which is time-consuming and far from automated rule interpretation. And manual interpretation based on DSQLs or function library demands high proficiency, posing challenges to automated rule interpretation. Third, function library are in their nascent stages and have only been explored in dealing with Korean building codes. Further research is necessary to explore the adaptability of similar approaches to building codes in other countries to facilitate automated rule interpretation. Finally, directly converting complex clauses into executable code through LLM-based atomic function matching remains largely unexplored. Therefore, this work presents LLM-FuncMapper, a method to match clauses to atomic functions and then automatically compose them into executable code based on the LLMs.

# 3 Methodology

The proposed LLM-FuncMapper comprises two key components: the atomic function library to capture shared computational logics of implicit properties and complex constraints, and the rule-based adaptive prompt engineering for function matching and code generation based on LLM. Then the effectiveness of the proposed methods is validated using several statistical analyses, experiments, and proof of concept, as illustrated in Fig.1.

The construction of the atomic function library aims to identify and explicitly define atomic functions within the clauses, thereby representing and expressing the complex computational logic and serving as the foundational blocks for interpreting regulatory clauses (Section 3.1).

The rule-based adaptive prompt engineering for LLM-based function matching aims to recommend the most relevant atomic functions from the library for each clause to be interpreted, reducing the time and effort required for domain experts to find the most suitable functions. To realize this, several LLMs are evaluated, a prompt template with the chain of thought (CoT) is designed, and the rule-based adaptive strategy is proposed, illustrated in Section 3.2.

In the validation stage, descriptive and statistical analyses are conducted to validate the expressibility of the function library (Section 4). Then, the performances of the proposed LLM-based atomic function matching method are thoroughly examined via experiments (Section 5). Finally, a rule checking of an actual plant is conducted as a proof of concept to validate the capability of converting complex regulatory clauses into executable code through the composition of atomic functions (Section 6).

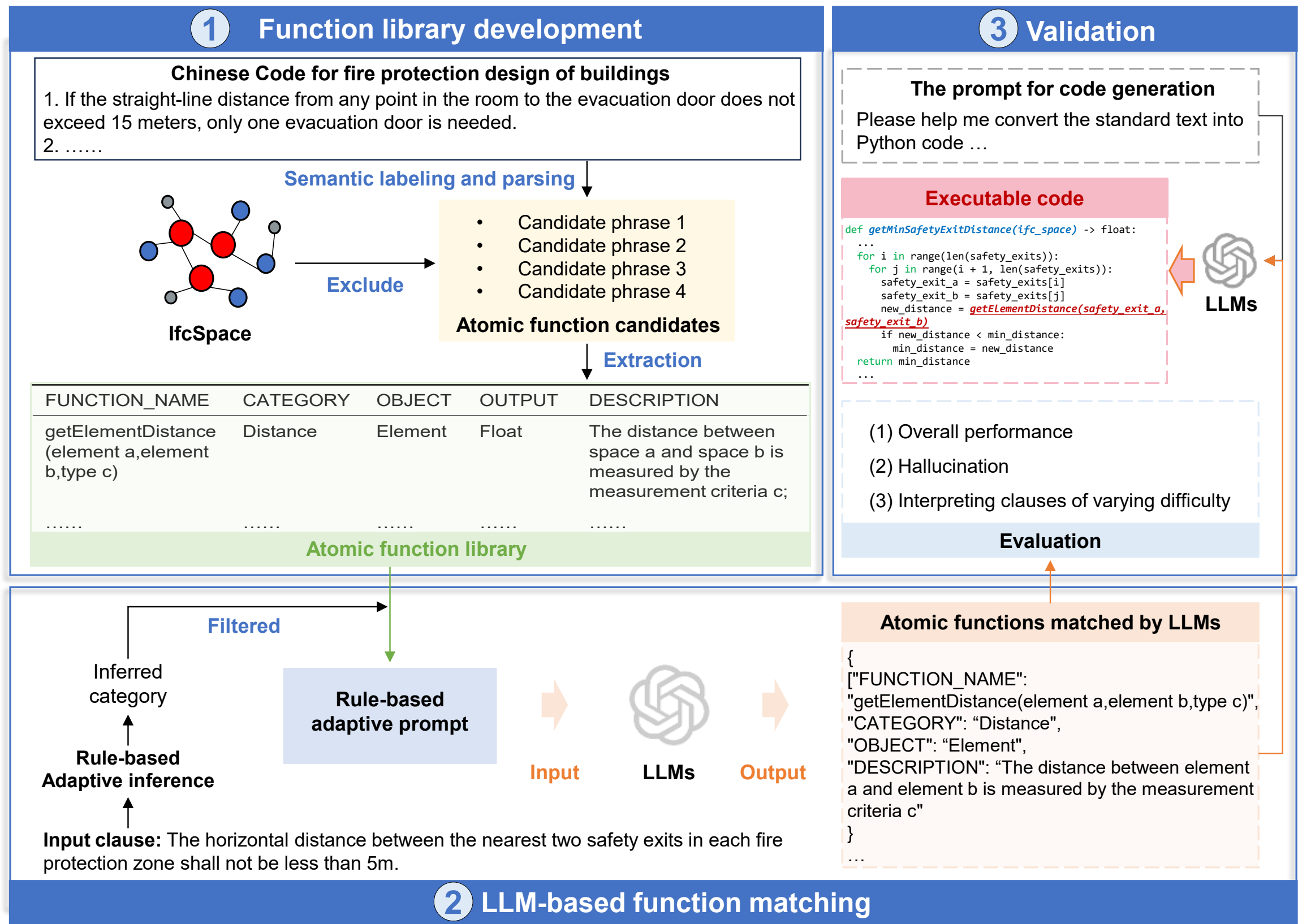

Fig. 1 The workflow of this study

## 3.1 Development of atomic function library

### 3.1.1 Data acquisition and preprocessing

**Step 1: data acquisition**

For the purpose of this research, Chapters 3, 4, and 5 of the *Chinese Code for Fire Protection Design of Buildings* (GB 50016-2014), one of the most widely used codes in China, have been chosen. These chapters cover regulations related to warehouses and plants (Chapter 3), storage areas and combustible material storage areas (Chapter 4), and civil buildings (Chapter 5). The selected clauses within these chapters encompass some complex clauses, such as building fire resistance rating, the distance of fire separation, evacuation distance, and plane layout. These complex clauses pose challenges in terms of direct description using the first-order logic. Detailed information on the collected data is illustrated in Section 4.1.

**Step 2: data preprocessing**

Preprocessing includes sentence splitting, tabular clauses converted into textual clauses, and non-computer-processable clause filtering. Sentence splitting aims to break down long clauses that contain multiple design requirements into short clauses that contain a single design requirement. After splitting, each clause contains complete elements for checking, including the objects to be checked, the required constraints and conditions for checking, and the specific requirements of the objects. Then, tabular clauses are converted into the textual format that is expressed in natural language. Before the tables, some clauses typically define the unified or similar objects to be checked. Then, the table contents contain specific requirements under different conditions. Hence, it is necessary to combine the description before the table and the requirements in the table to form a short sentence with a complete structure that enables a computer to interpret it. After clause splitting and conversion, the non-computer-processable clause filtering aims to identify and filter out sentences that are not easily interpreted by a computer. Non-computer-processable sentences are those that require additional guidelines for a machine to determine whether they are "satisfied" or "failed" (Uhm et al., 2015). Only computer-processable sentences are retained for further analysis. The non-computer-processable clauses primarily fall into the following three types:

(1) Definitions clauses. These clauses do not have requirements on objects but serve to introduce the definition or category of the objects. E.g., The width of evacuation walkways, stairs, doors, and safety exits in theaters, cinemas, auditoriums, gymnasiums, and other places shall comply with relevant regulations (in Chinese).

(2) Qualitative clauses that require subjective judgment. The requirements contained in these clauses have vague words and lack clear standards. It needs to be evaluated and analyzed by domain experts to confirm whether the design is satisfactory or not; E.g., placing the civil buildings near the factory buildings is not recommended (in Chinese). "Near" here lacks a clear definition and is difficult for a computer to check.

(3) Clauses with external references. E.g., Other fire protection designs should comply with the Chinese code for fire protection design of thermal power plants and substations (GB 50229) and other standards. (in Chinese).

The preprocessing is manually completed by domain experts to ensure that the meaning of the split sentences and converted tables is consistent with that of the original sentences and tables, and to ensure that the non-computer-processable clauses are filtered out.

### 3.1.2 Function extraction and library establishment

After data preprocessing, we adopt the NLP-CFG-based semantic labeling and parsing method (Zhou et al., 2022; Zheng et al., 2022a) to convert clauses into a syntax tree format that explicitly exposes objects, attributes, and conditions, facilitating sentence analysis and atomic function extraction (Fig. 2). We then construct the atomic function library by determining the functions needed to extract and compute all identified phrases, ensuring that the objects and attributes mentioned in the regulations can be extracted using this method. To reduce manual library building effort, phrases are matched against the IFC schema, which is widely adopted for collaboration and model storage in BIM projects (IFCwiki, 2018). Matched phrases are filtered out because IFC already contains the relevant concepts or attributes, eliminating the need to create additional functions for those properties. Unmatched phrases, on the other hand, are retained as candidates for further analysis and atomic function extraction. This significantly streamlines the process and avoids unnecessary duplication. While IFC continues to expand, the proposed method remains valuable because IFC cannot incorporate all project-specific or specialized regulations, and the same concept-matching strategy applies to non-IFC data formats as well.

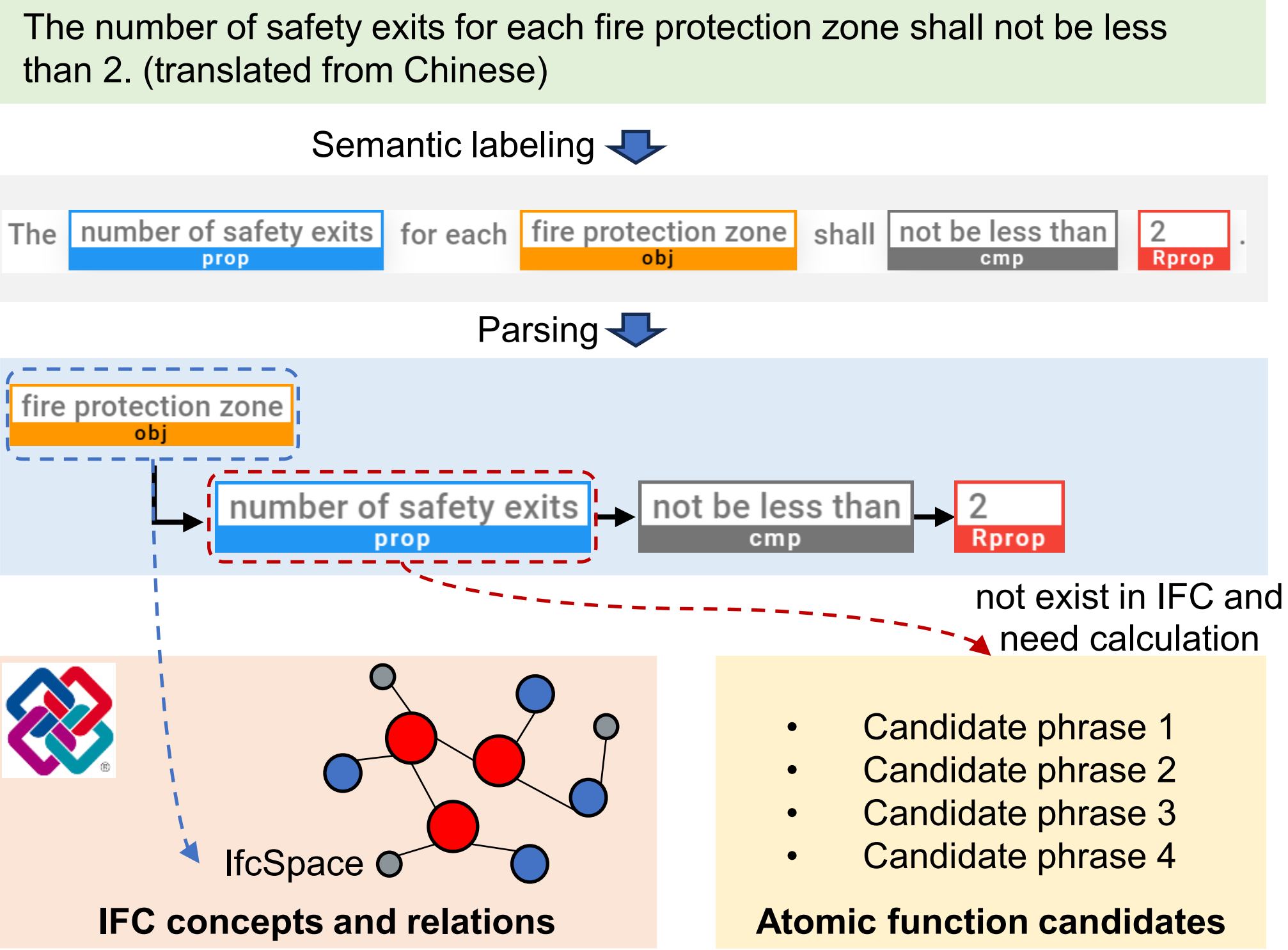

Fig. 2 Using the semantic labeling and parsing tool to assist atomic function extraction

The extraction of atomic functions hinges on determining the function's inputs and outputs based on the candidate phrases. Inputs are the attributes explicitly stored on the constrained object; outputs are the computed results derived from those inputs. Because object types differ in attribute sets and representations (e.g., IfcSpace vs. IfcBeam), overly general inputs hinder implementation. We therefore classify inputs by grouping objects with similar attributes into categories before defining function signatures.

Thus, establishing the atomic functions library involves two essential steps: (1) Categorize the objects to be checked, aiming to identify the constrained objects in the clauses, and subsequently merge and summarize their categories. These objects serve as the parameters of atomic functions; (2) Atomic function extraction aims to match clauses to predefined atomic functions. Then, we define and record the unique functions to form the function library.

**Step 1: classifying the objects to be checked**

The objects are categorized into five categories, including building, space, element, system and equipment, and goods, considering the compatibility with IFC (ISO 16739-1: 2018) and Standard for building information modeling semantic data dictionary (building fascicule) (SJGXXX-2023).

The terms building, space, and element are equivalent to ifcBuilding, ifcSpace, and ifcBuildingElement in IFC, respectively. However, IFC does not have a single term that defines the various mechanical and electrical equipment, furniture, and systems (e.g., sprinkler systems) in buildings. Therefore, this study utilized the term system and equipment to define the concepts of equipment, furniture, and systems with reference to SJGXXX-2023 (SJGXXX-2023). Furthermore, considering the clauses related to factory buildings often involve sources and materials (such as combustible gases and liquids), the term goods is introduced to capture such concepts. Table 3 of Section 4 summarizes the distribution of the object types in the GB 50016-2014.

**Step 2: extraction of atomic functions for rule checking**

Due to the inherent ambiguity of natural language, extracting functions is not as straightforward as the extraction of objects, often requiring the interpretation and intervention of domain experts. We analyze the phrases that constrain objects and make their shared computational logic explicit. Functions are named in camelCase with a leading verb and a descriptive noun/adjective. Verbs fall into three types: "get," "is," and "has". Functions starting with "get" return collections of objects, strings, or numeric values. Functions beginning with "is" and most functions starting with "has" return boolean values. The parameters of a function include (1) obj, the objects to be checked, and (2) type: the method by which the checking process is conducted. A typical example is *Float getSpaceDistance(space a, space b, type c)*. This function calculates the distance between two spaces and returns a float value. The parameters of the function include two space objects (a and b) and a type parameter. The type indicates the method used to measure the distance. Examples of distance measurement methods include linear distance, evacuation distance, and so on.

During the extraction process, the involved functions can be categorized into low-order and high-order types. Low-order functions that can only complete the checks based on a single or a small number of explicit model data and can be expressed in first-order logic, but they cannot handle complex checks. While the high-order functions can derive implicit properties from explicit data and thus can handle phrases with complex computational logics that low-order functions cannot capture.

**Step 3: information enrichment**

This step aims to record the unique functions extracted in step 2 to establish the library. For clarity, we categorize the functions into eight types based on their usage, including property, space_location, existence, quantity, geometry, distance, wall-window ratio, and area. Additionally, descriptions of the application scenarios for each function are also provided. Finally, the resulting record information includes the function category (i.e, "CATEGORY"), the input object category (i.e., "OBJECT"), the output type (i.e., "OUTPUT"), the function name (i.e., "FUNCTION_NAME"), and the meaning of the function (i.e., "DESCRIPTION"). A typical example is shown in Fig.3 below. A comprehensive overview of the atomic function library is shown in Table A (APPENDIX)

| CATEGORY | OBJECT | OUTPUT | FUNCTION_NAME | DESCRIPTION |
|---|---|---|---|---|
| Distance | Space | Float | getSpaceDistance (space a, space b, type c) | The distance between space a and space b is measured by the measurement criteria c; |
| …… | …… | …… | …… | …… |

Fig. 3 Example of the function recorded in the library

## 3.2 Rule-based adaptive prompting of LLM for function matching

### 3.2.1 Injecting domain knowledge with prompt engineering

Although LLMs trained on vast corpora excel at dialogue, reasoning, and program synthesis (Patil et al., 2023), they still struggle with specialized domain knowledge (Saka et al., 2023). When encountering the atomic function library defined in this work, beyond the extent of LLMs' pretraining datasets, the model cannot reliably map regulatory clauses to the required complex functions without additional guidance.

Domain knowledge can be injected via fine-tuning (Wang et al., 2023) or prompt engineering (Zuccon & Koopman, 2023). Because fine-tuning large models is resource-intensive, we adopt a prompt-engineering approach. Here, a "prompt" is a concise set of instructions that defines context, highlights salient information, and specifies the desired output form and content (White et al., 2023). Prompt engineering refers to the process of designing and refining the way you give instructions or questions (the "prompt") to a generative AI model. The goal is to get the most useful or accurate response from the model.

Injecting domain knowledge with prompt engineering can be formalized as follows. Suppose we have a knowledge base $a$ and a set of $k$ examples $\{(x_i, y_i)\}_{i=1}^{k}$, that are provided as part of the test-time input (Gao et al., 2022), where $k$ is usually a very small number, $x_i$ is the input text, and $y_i$ is the corresponding output retrieved from the given knowledge base $a$. Then the prompt $p = a \,||\, \langle x_1 \cdot y_1 \rangle \,||\, \ldots \,||\, \langle x_k \cdot y_k \rangle$, where "$\cdot$" means the concatenation of the input and output of each example, and "$||$" means the concatenation of the knowledge base and different examples. During the inference stage, the target instance $x_{text}$ is appended to the prompt $p$, and $p \,||\, x_{test}$ is passed to the LLMs to generate the answer $y_{test}$. Note that the prompt does not require back-propagation and modify the parameters of LLMs.

### 3.2.2 Prompt template design with chain of thought

In this work, to enable LLMs to understand the domain-specific knowledge contained in the proposed atomic function library and further match clauses to proper functions based on requirements, we design a prompt template, as depicted in Fig. 4. The proposed prompt template consists of four parts: role, library, example, and an analysis section.

(1) The role part defines the role and goal that LLM needs to fulfill and accomplish, which is to serve as an API recommendation system and match each input clause to the most suitable functions from the atomic function library. In this study, the model is instructed to select up to 5 atomic functions per clause, reflecting evidence from our prior work on building-code atomic functions (Lu et al., 2023) that indicates clauses typically contain no more than five. In practical applications, this number can be flexibly adjusted according to the complexity of the target norms, allowing the model to handle clauses with a larger set of functions when necessary.

(2) The prompt of the library part serves to inject the information from the developed atomic function library into the LLM, enabling it to become acquainted with the relevant details of the atomic function library. Specifically, the prompt of the library part describes the following information, including the function category (i.e, "CATEGORY"), object category (i.e, "OBJECT"), function name (i.e, "FUNCTION_NAME"), the meaning of the function (i.e, "DESCRIPTION"), and natural language phrases from codes that employ the function (i.e., "EXAMPLE"). The provided few examples aim to allow the LLMs to better understand the semantics of the function, which is called few shot prompting strategy (Logan et al., 2021).

"goal":"You are a helpful API recommendation. You should recommend some APIs for the sentence, based on the requirements and the provided $API_LIBRARY, to support translate the sentence into code.",

(1) Goal part defines the tasks

Here is the $API_ LIBRARY:
{
["**CATEGORY**": "area",
"**OBJECT**": "space",
"**FUNCTION_NAME**": "getFloorArea(space a)",
"**DESCRIPTION**": "get the area of space a",
"**EXAMPLE**": ['Floor area of the room', ……],
}

(2) Library part injects the information of high-level functions

"requirements": "Choose at most 5 APIs from the above provided $API_LIBRARY for the sentence. You first determine the category, the object, and then the function_name **step_by_step**. Do not repeat the format in your answer.
The answer should follow the format:
{
<<<category>>>: $CATEGORY,
<<<object>>>: $OBJECT,
<<<function_name>>>: $FUNCTION_NAME,
<<<description>>>: $DESCRIPTION,
}",

(3) Requirement part defines the detailed output format

**Zero shot chain of thought**

Here are some examples for you:
[1] example 1, sentence: "The maximum area of the fire compartment cannot be greater than 2000"
Inference steps:
The category inferred from "maximum area" is area, the object inferred from "fire compartment" is space then the funtion_name should be getFloorArea(space a,type b).
Thus, the first answer is:
{
"category": area,
"object": space,
"function_name": getFloorArea(space a,type b),
"description": get the area of space a via type b
}

(4) Example part informs the thinking process (**Chain of thought**)

Here is the sentence:

(5) Analysis part inputs the rule to be analyzed

Fig. 4 Prompt for atomic function matching

Notably, all prompts fed into the LLMs on the left side of the figure in practical applications are provided in Chinese.

(3) The example part aims to provide several random analyses of cases that guide the LLM's

thought process, resembling human beings' reasoning process step by step, through the use of the CoT technology (Wei et al., 2022). The main idea of CoT is that by showing the LLM some few shot exemplars where the reasoning process is explained in the exemplars (Wei et al., 2022). In this particular task, the LLM is first tasked with identifying the key phrases related to atomic functions, then inferring the function category and object category based on the key phrases, and ultimately providing the proper functions accordingly.

(4) Lastly, the analysis part pertains to the clause that needs to be analyzed.

### 3.2.3 Rule-based adaptive prompt tuning

Furthermore, when dealing with a large number of atomic functions within the atomic function library, recommending functions becomes more challenging for the LLM. Besides, because the prompt in the library part is too long, the length of the whole prompt may exceed the maximum token of LLMs. To address this, we propose a rule-based adaptive prompt tuning method to infer the function categories of the input clauses and further refine the prompt before utilizing the LLM for function matching. The purpose of inference is to filter out the atomic functions that are completely irrelevant to the given clause, thereby reducing the selection scope for the LLM and refining and shortening the prompt, as illustrated in Fig. 5.

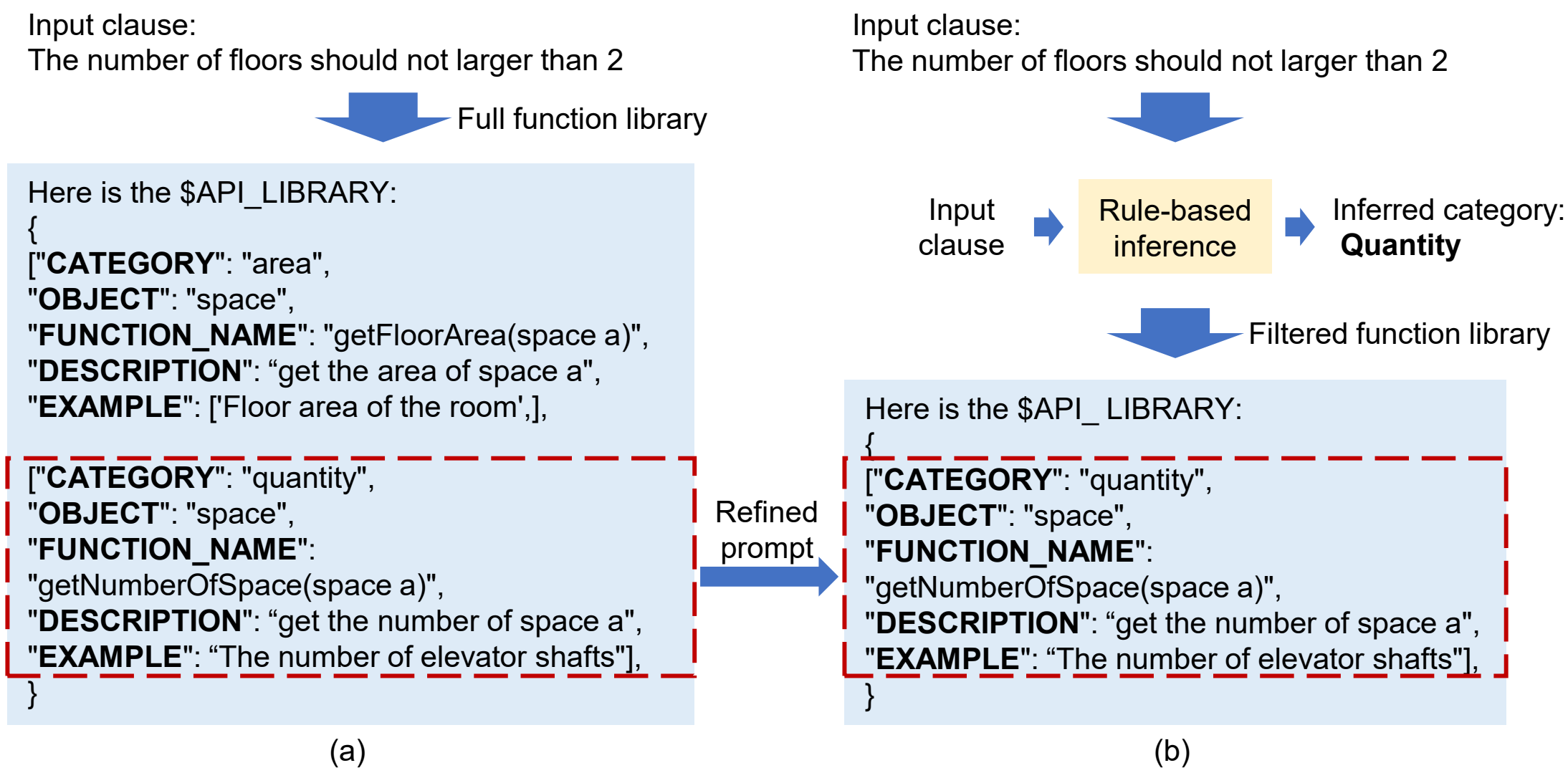

Fig. 5 Rule-based adaptive prompt tuning method (a) the full prompt, (b) the rule-based adaptive prompt

In this study, we adopt a rule-based inference method because ML alternatives require large labeled datasets. To implement this method, we define a table of keywords for inference, with a subset presented in Table 2, and the complete list is in Appendix Table B. When a clause contains a keyword from a given category, we infer that its atomic functions may belong to that category; a clause may span multiple categories. It is important to note that this process focuses only on the categories of quantity, geometry, distance, and area, while the other categories are implicitly included in the prompt. This decision is based on three main reasons. First, almost all clauses contain the property category. Second, the keywords related to the space_location category and the existence category exhibit a high overlap rate, making it difficult to distinguish between them. Such ambiguity could potentially lead to function-matching errors caused by inference errors. Third, very few clauses use the functions of the wall_window_ratio category, so this category is not considered in the inference. Section 5 reports experiments validating this procedure.

Table 2 Keywords for rule categories

| Category | Keywords (in Chinese) |
| --- | --- |
| quantity | Number, times… |
| geometry | Length, width, height, higher than… |
| distance | Distance, distance bewteen… |
| area | Area, volume… |

## 3.3 Validation

The validation stage aims to validate the expressibility of the function library for rules with complex computational logic and the performance of the proposed rule-based adaptive prompt engineering for function matching based on LLM. Thus, descriptive and statistical analyses are conducted for the validation of the expressibility of the function library (Section 4). Then, we evaluated the performance of the proposed LLM-based atomic function matching method (Section 5.1) using the experimental settings (Section 3.3.1) and metrics (Section 3.3.2) described in the subsections. Finally, a rule checking of an actual plant is conducted as a proof of concept to validate the effectiveness of the proposed method in ARC (Section 5.2).

**3.3.1 Experiment settings**

This section delineates the experiments conducted to authenticate the efficacy of our proposed method. The dataset for experiments encompassed 410 sentences randomly selected from the annotated dataset, as described in Section 4.1, along with their corresponding categories and atomic functions.

First, we evaluated the rule-based inference to ensure that it provides sufficient atomic functions for the LLM while minimizing the number of input tokens.

Then, we evaluated the performance of several LLMs on the function matching task. The models tested included ChatGPT4o (OpenAI, 2024), Claude 3.5-Sonnet (Anthropic, 2024), Deepseek-V3 (DeepSeek, 2024), and Deepseek-R1 (DeepSeek, 2025), which were among the most advanced LLMs available at the time. ChatGPT4o are primarily based on OpenAI's GPT architecture, utilizing the transformer structure, and achieving performance through large-scale data pretraining and supervised fine-tuning. Claude 3.5-Sonnet, similarly based on the transformer architecture, allows users to more easily control the style and direction of responses, catering to different conversational needs. Deepseek-V3 and Deepseek-R1, as emerging models primarily trained in Chinese, prioritize real-time adaptation to dynamic conversation patterns, making them particularly effective in specialized applications such as technical support and personalized customer service. Overall, ChatGPT4o, Claude 3.5-Sonnet, Deepseek-V3, and Deepseek-R1 demonstrate superior capabilities in language understanding, conversation generation, and multi-task handling, particularly in their ability to manage complex contexts and deliver high-quality, contextually relevant responses, which is the primary reason for our selection of these four models. These models were trained on multilingual data, giving them strong multilingual capabilities, including robust support for high-resource languages like Chinese (Enis & Hopkins, 2024; Bang et al., 2023; Robinson et al., 2023). By using Chinese prompts, these models can effectively interpret Chinese clauses. For broader accessibility, we translated this workflow into English in the manuscript, as shown in Fig. 4. It is important to note that these models could be replaced with more advanced LLMs in the future to achieve higher precision in function matching.

Specifically, we input the full prompt into LLMs (i.e., following the approach depicted in Fig. 5 (a)) to identify the atomic functions used in the 410 sentences. Then, we employed the rule-based inference strategy to acquire the refined prompt and predicted the atomic functions of each clause using the refined prompt (i.e., following the approach depicted in Fig. 5 (b)). Besides, we compared the performance of the fine-tuned BERT model. We finetuned the BERT model for a multi-label classification task (HuggingFace, 2023), enabling it to output the five most probable functions for each sentence. The fine-tuning process utilized 918 sentences as the training dataset and 164 as the validation dataset, ensuring no overlap with the 410 sentences in the previous test dataset. The training dataset was employed for training and updating the BERT model, while the validation dataset was used to test the performance of the model and choose the best combination of hyperparameters and the best model. The test dataset is used for final evaluation. The odels underwent fine-tuning with a maximum of 20 epochs, surpassing the epoch requirement for the optimal model. Furthermore, the impacts of various learning rates and batch sizes were examined through a grid search, considering learning rates of $1\times10^{-4}$, $1\times10^{-5}$, and $1\times10^{-6}$, and batch sizes of 2, 4, 6, 8, and 16. The optimizer is AdamW.

Additionally, the occurrence of hallucinations in LLMs was evaluated using the hallucination rate. In this study, hallucinations are defined as instances where the LLMs suggest functions that do not exist in the function library, referred to as 'fake functions'. For instance, as illustrated in Fig. 6, while the LLM accurately predicted three atomic functions, it also fabricated a function not found in the library, which is highlighted in red. A higher rate of hallucinations implies a greater number of recommended fake functions, leading to potential confusion.

**Clause** When an atrium is incorporated within a building, if the fire compartmentation area exceeds that specified in Section 5.3.1 of this regulation, no combustible materials should be placed within the atrium.
(in Chinese: 当建筑内设置中庭，其防火分区面积大于本规范第5.3.1条时，中庭内不应布置可燃物。)

| | Function1 | Function2 | Function3 | Function4 | Function5 |
|---|---|---|---|---|---|
| **Ground truth** | existence: hasSpace(building a,space b) | area: getFloorArea(space a,type b) | existence: hasGoods(space a, goods b) | property: getProperty(goods a,property b) | |
| **Identified functions** | getSpaceLocation(space a); ✗ | area: getFloorArea(space a,type b) ✓ | existence: hasGoods(space a, goods b) ✓ | property: getProperty(goods a,property b) ✓ | property: getProperty(space a,property b); ✗ |

category
function name
object type
fake function
✗ predict false
✓ predict true

Fig. 6 Exemplary outputs of hallucinations in LLMs

Finally, to further investigate the applicability of LLMs in interpreting regulations of varying difficulty, their performance for simple clauses and complex clauses involving different numbers of atomic functions was contrasted. It is important to note that simple clauses contain only the functions of the property category (low-order functions), while clauses containing any other functions are considered complex clauses. For a more detailed analysis, we further categorized the complexity of complex clauses based on the number of atomic functions they involve.

### 3.3.2 Validation metrics for function matching

To measure the rule-based inference result, we use the recall rate as the most crucial performance indicator in our study:

$$Recall = TP\ /\ (TP\ +\ FN) \tag{1}$$

where $TP, FP, TN,$ and $FN$ denote the number of *True positive, False positive, True negative,* and *False negative* for each label.

This is because a high recall rate means the algorithm can find as many target categories as possible. In this case, the prompt can be designed to encompass all atomic functions for the target categories, even with some redundancy, allowing the LLMs to select the correct functions and thereby ensuring the accuracy of the final outcome. Given that there will be a subsequent manual selection of functions, any unrelated functions can be manually filtered out.

To measure the applicability of the proposed LLM-based atomic function matching method, the results predicted by models are compared with the gold standard, and the widely used metrics for recommender systems Recall@k (Malaeb, 2017; Melchiorre et al., 2020) are calculated as follows:

$$Recall@k = Number_{pred@k}\ /\ Number_{true} \tag{2}$$

$$Recall@k_{avg} = (\sum_{i}^{n} Recall@k_i)/n \tag{3}$$

where $Number_{pred@k}$ denotes the number of relevant atomic functions among the top $k$ matched functions for each clause. $Number_{true}$ denotes the number of all relevant functions for each clause. $Recall@k_{avg}$ denotes the average $Recall@k$ value of all clauses. In this work, $k$ is set to 5 according to previous studies (Xia et al., 2013; Wang et al., 2018).

Finally, to evaluate the occurrence of hallucinations in LLMs, we calculated the Hallucination rate:

$$\text{hallucination rate} = \text{Number of FF} / \text{Number of RF} \tag{4}$$

where "Number of FF" is short for "Number of fake functions". "Number of RF" is short for "Number of recommended functions".

# 4 Developed function library

## 4.1 Results of data acquisition and preprocessing

Regulatory texts from Chapters 3, 4, and 5 of the *Chinese Code for Fire Protection Design of Buildings* (GB 50016-2014) are utilized as the data source for this study, a widely recognized standard in the field of fire protection, which to some extent reflects the typical language style of Chinese regulations. A total of 451 textual clauses were collected. Then the long clauses were split into short clauses, resulting in a total of 679 textual clauses. Subsequently, a total of 1048 tabular clauses are then converted into the textual format, bringing the cumulative count to 1727. After filtering out non-computer-processable clauses, 235 clauses were excluded. As a result, a total of 1492 computer-processable clauses remained and were suitable for subsequent analysis. Two experts with experience in rule interpretation were then tasked to identify sentence categories and atomic functions encapsulated within these clauses.

## 4.2 Descriptive analysis of objects involved and atomic functions

Table 3 summarizes the distribution of types of objects in Chapters 3, 4, and 5 of the *Chinese Code for fire protection design of buildings* (GB 50016-2014). In Chapter 3, most design clauses focus on the object in the building category (45.86%), followed by objects in the space category (29.17%). As for Chapter 4, the design clauses are also mainly concentrated on the objects in the building category (48.35%), similar to that of Chapter 3. However, in Chapter 4, there is a higher proportion of clauses related to the objects in system and equipment category, accounting for 30.47%. Compared with the former two chapters, Chapter 5 focuses on the check of objects in the element category and the space category, accounting for 37.51% and 30.57%, respectively.

Table 3 Distribution of object types

| | space | | element | | equipment | | building | | goods | |
|---|---|---|---|---|---|---|---|---|---|---|
| | frequency | % | frequency | % | frequency | % | frequency | % | frequency | % |
| Chapter 3 | 528 | 29.17% | 256 | 14.14% | 48 | 2.65% | 830 | **45.86%** | 148 | 8.18% |
| Chapter 4 | 201 | 13.82% | 39 | 2.68% | 443 | 30.47% | 703 | **48.35%** | 68 | 4.68% |
| Chapter 5 | 568 | 30.57% | 697 | **37.51%** | 128 | 6.89% | 451 | 24.27% | 14 | 0.75% |

Table A (in APPENDIX) presents a comprehensive overview of the atomic function library established in this study, encompassing 66 atomic functions. The functions are classified into eight categories based on their usage, including existence, quantity, geometry, distance, area, space location, window-to-wall ratio, and property. Table A introduces the category, type of return values, function name, and the corresponding description of each atomic function. The atomic function library consists of 5 low-order functions and 61 high-order functions, with the low-order functions distinguished by being highlighted in green. The low-order functions primarily consist of functions within the property category, as they can only process a single or a small number of explicit data from the model. Functions in the other categories involve more intricate computational logics, making them high-order functions.

## 4.3 Statistical analysis of atomic functions

Then, this study counts the occurrence of defined functions in different clauses. Fig. 7 illustrates the frequency distribution of the functions that have been utilized more than 10 times. Notably, low-order functions are represented in pink, while high-order functions are denoted in blue. The analysis reveals a long-tail distribution in function usage, suggesting that algorithm engineers should prioritize the implementation of functions that are frequently used. Among the top five most utilized functions, three functions are low-order functions (i.e., *getProperty(building a, type b)*, *getProperty(space a, type b)*, *and getProperty(element a, type b)*). The remaining two functions are high-order functions (i.e., *getFireproofDistance(building a, building b, type c)* and *hasSpace(building a, space b)*).

Notably, the high-order functions that are most commonly employed include *getFireproofDistance(building a, building b, type c)*, *hasSpace(building a, space b)*, *hasElement(building a, element b)*, *hasElement(Space a, element b)*, and *hasEquipment(building a, equipment b)*. This observation highlights that high-order functions are mostly used in checking fire separation distances between buildings and extracting the containment relationships among objects.

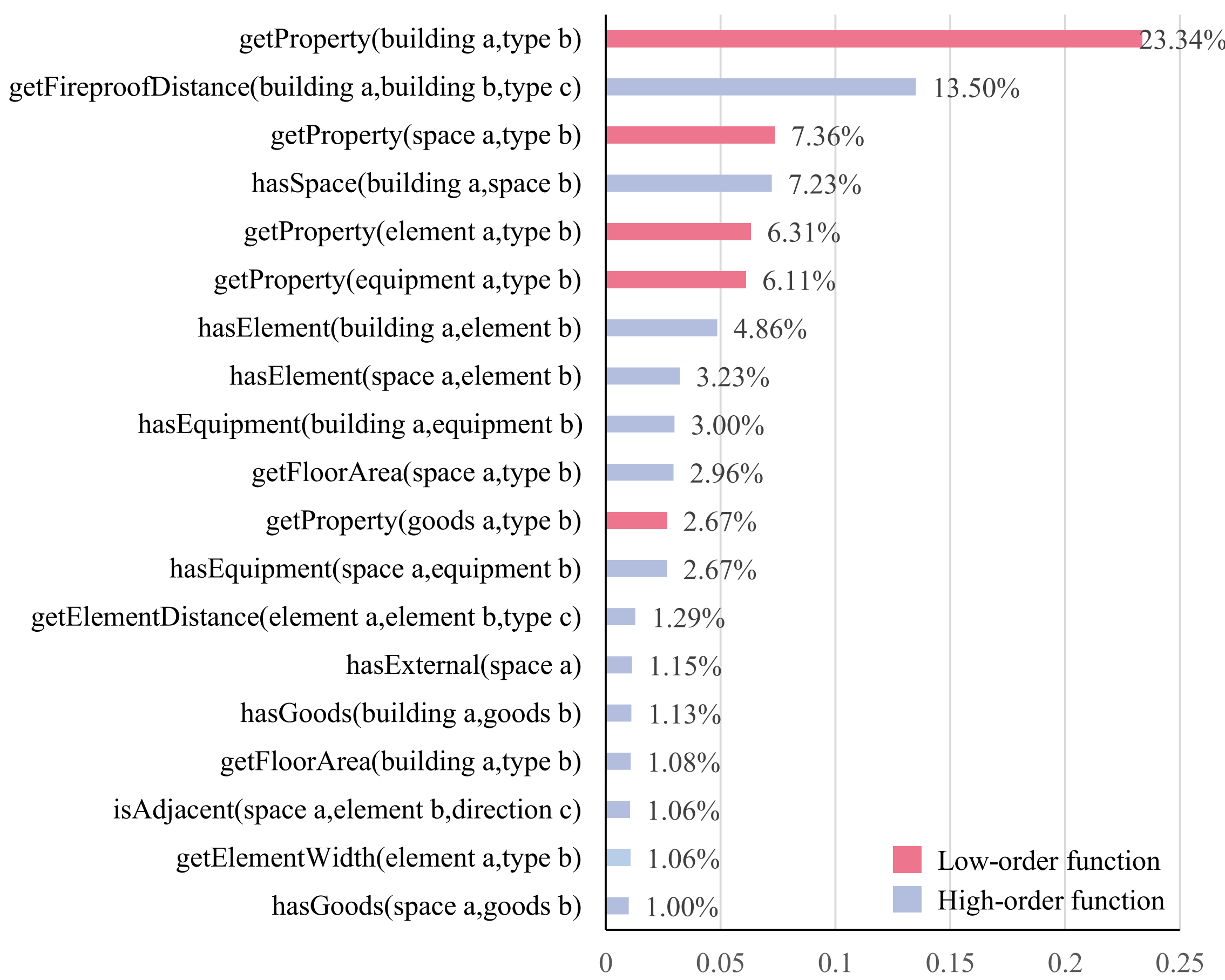

Fig. 7 Usage distribution of partial functions among 66 functions

Additionally, Fig. 8 provides insights into the usage proportion of functions of different categories. It is worth noting that the functions of the property category (low-order functions) exhibit the highest frequency of usage. This is attributed to the frequent need for retrieving properties associated with the objects to be checked, making it the most commonly used function. Following the property category, the second most utilized category is the existence category. This prominence stems from the frequent need to inquire about the containment relationship among objects during the checking of the clause from fire codes. For instance, clauses such as "The number of safety exits for each fire protection zone shall not be less than 2" involve the containment relationship between the fire protection zone and the safety exits. The third most frequently employed category is the distance category. This is primarily due to the checking of fire protection distances and evacuation distances that are often involved in the checking of the fire code.

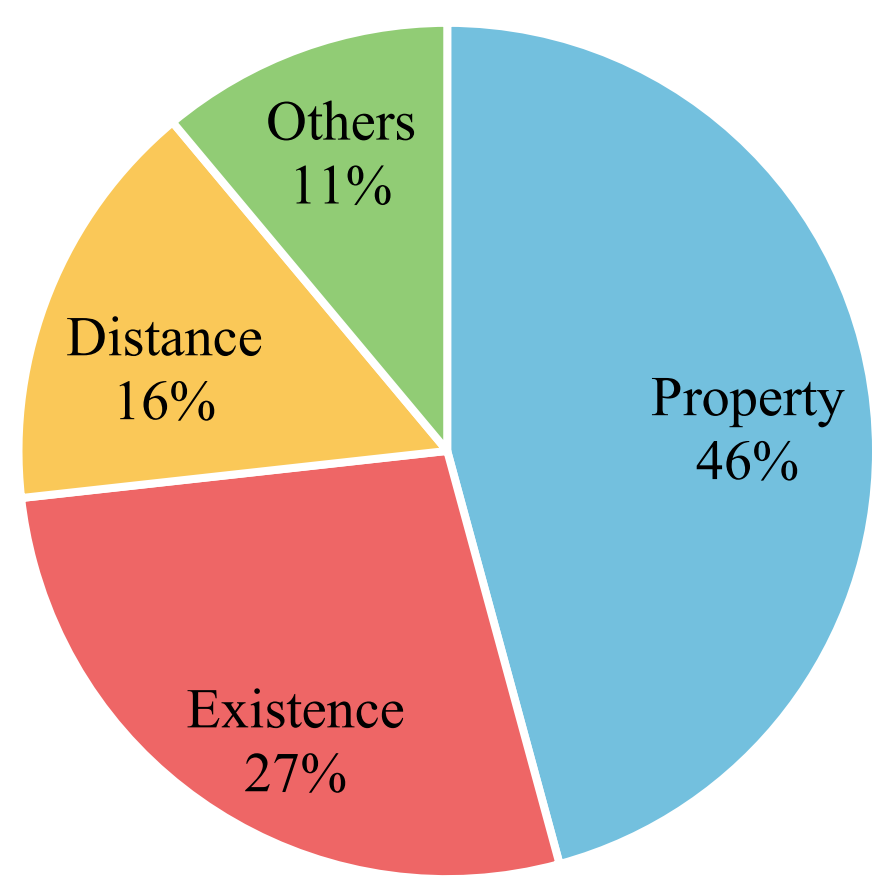

Fig. 8 Usage proportion of functions of different categories

# 5 Results

## 5.1 Performance of the proposed rule-based adaptive prompt method

Table 4 presents the results of the rule-based adaptive prompt method proposed in this study. It demonstrates that satisfactory recall values of all categories (i.e., Recall values of all categories are higher than 92%) are achieved by the adopted method. This indicates that the proposed approach can provide sufficient atomic functions for the LLM while minimizing the number of input tokens, thereby improving model efficiency, relevance, and response quality by focusing on key information and reducing unnecessary complexity.

Table 4 Performance of the rule-based adaptive prompt method

| Category | Number | Recall |
|---|---|---|
| quantity | 27 | 100% |
| geometry | 44 | 100% |
| distance | 90 | 92.2% |
| area | 61 | 98.4% |

## 5.2 Overall results of atomic functions matched by different methods

Fig. 9 presents the overall results of atomic functions matched by different methods. Among the LLM-based approaches, Claude 3.5-Sonnet combined with a rule-based adaptive prompt achieved the highest recall@5avg value of 81.55%. This indicates that approximately four of the recommended five functions are the right functions needed to interpret the current clause. Notably, the performance achieved using Claude 3.5-Sonnet with a rule-based adaptive prompt surpasses that of the fine-tuned BERT model, which scored 68.5%, by 19%. This suggests that LLMs employing few-shot methods can outperform fine-tuning approaches, potentially eliminating the need for extensive manually annotated data for model fine-tuning, while also demonstrating the superiority of the method proposed in this study. Moreover, the results displayed in Fig. 9 also demonstrate that the rule-based adaptive prompt approach proposed in this study can significantly enhance the performance of LLMs. It increased the recall@5avg score by 4.6%, 5.6%, 1.2%, and 2.8% for ChatGPT4o, Claude 3.5-Sonnet, Deepseek-V3, and Deepseek-R1, respectively. This indicates that the rule-based adaptive prompt method proposed can substantially improve the performance of LLMs without the need for finetuning or pretraining the LLMs.

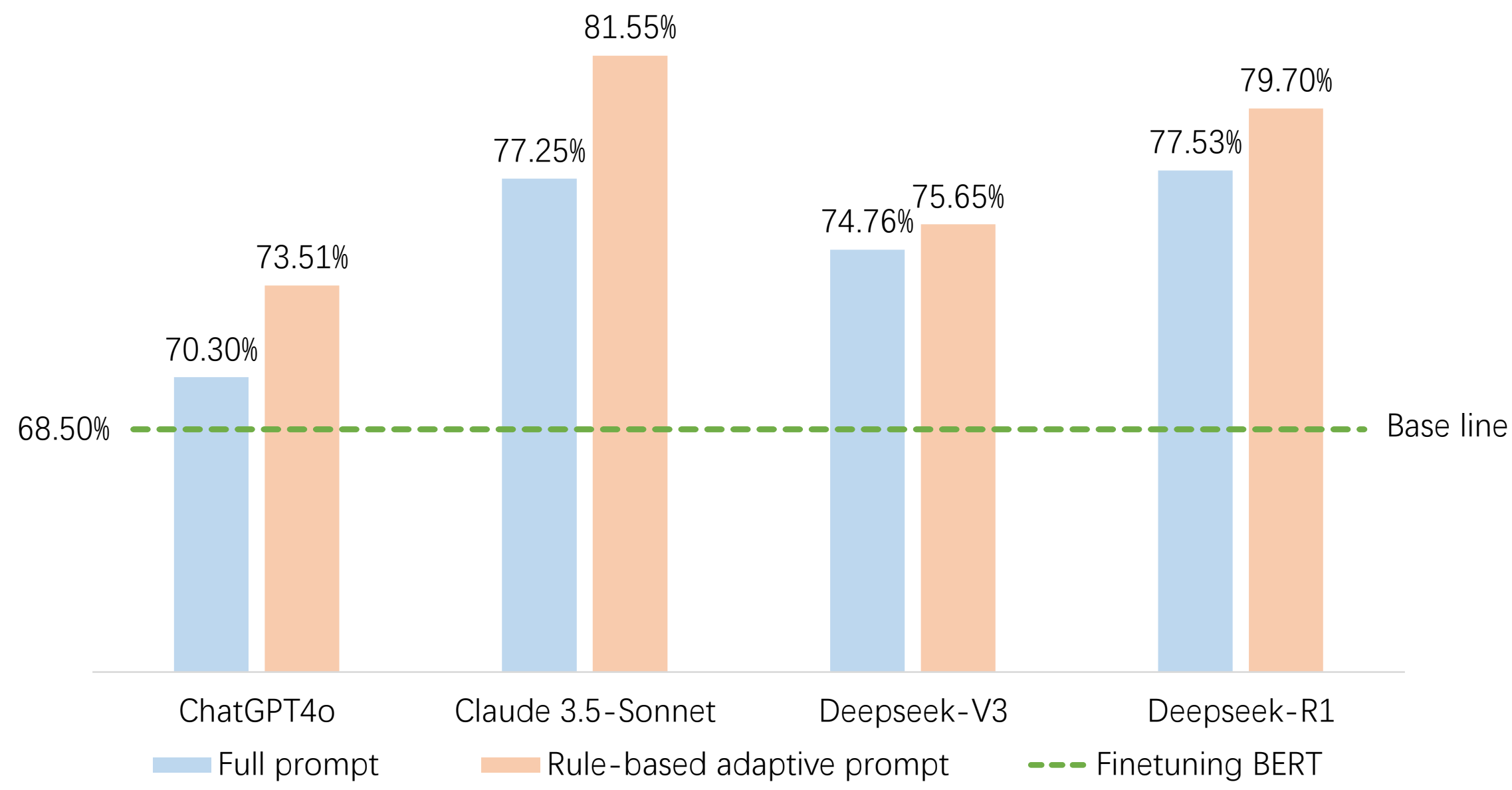

Figure 9 Performance of the function matching method (Recall@$5_{avg}$)

The typical case presented in Fig. 10 highlights the difference between the results obtained using the full prompt prediction and the rule-based adaptive prompt prediction. It can be seen that when using the full prompt, the LLM predicted an atomic function ("Function4" in Fig. 10) from the geometry category that is unrelated to the clause. While the ground truth categories of this clause are property and existence. In comparison to the ground truth, only two atomic functions were correctly matched, as shown in Fig. 10. Conversely, when using the rule-based adaptive prompt for prediction, the results did not contain functions from irrelevant categories, and all four atomic functions were predicted accurately, as shown in Fig. 10.

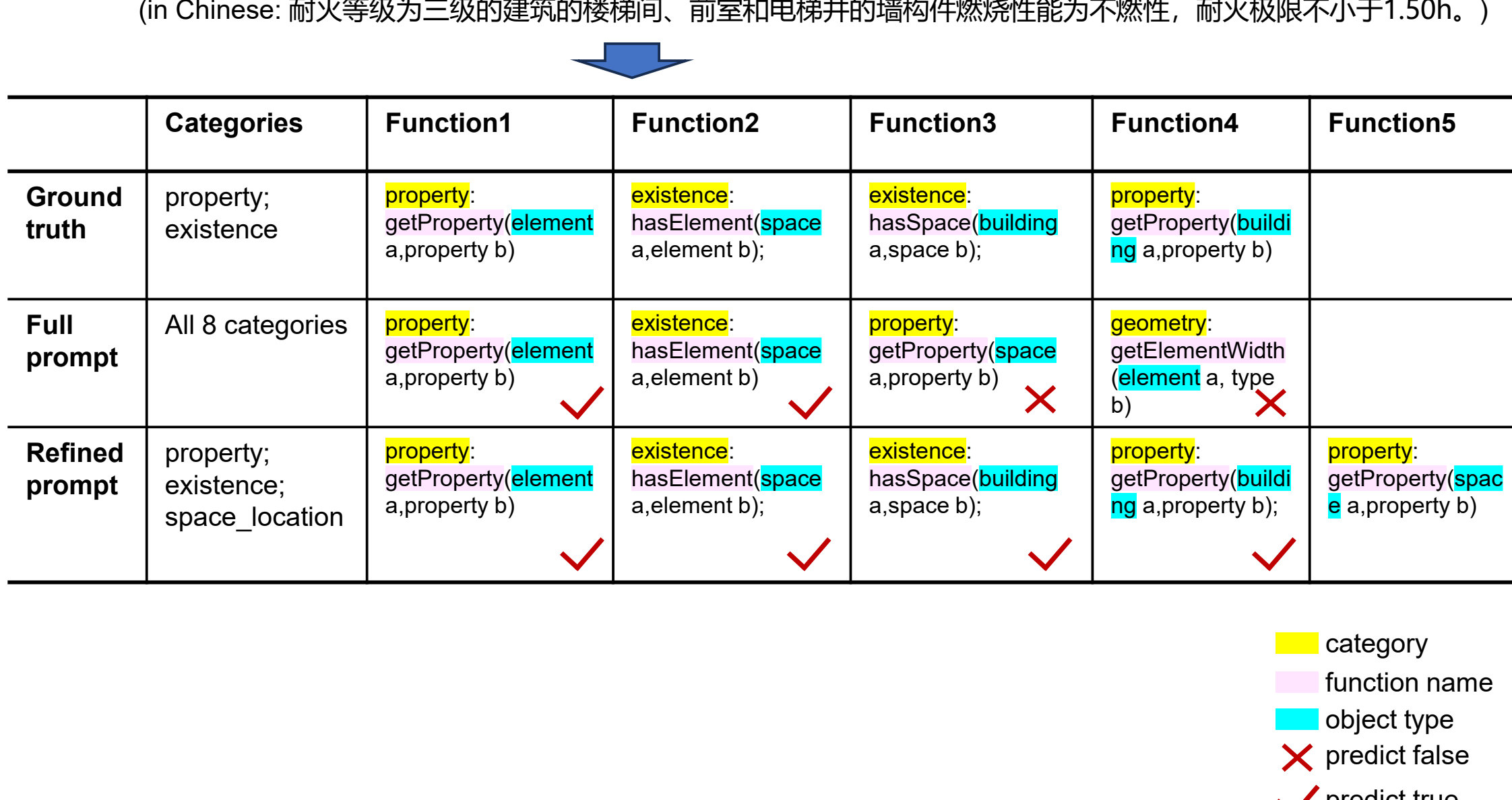

**Clause** When the fire resistance rating of the building is Class 3, the wall of the stairwell, front room and elevator shaft is non-combustible, and the fire resistance limit of the wall is not less than 1.50h.
(in Chinese: 耐火等级为三级的建筑的楼梯间、前室和电梯井的墙构件燃烧性能为不燃性，耐火极限不小于1.50h。)

| | Categories | Function1 | Function2 | Function3 | Function4 | Function5 |
|---|---|---|---|---|---|---|
| **Ground truth** | property; existence | property: getProperty(element a,property b) | existence: hasElement(space a,element b); | existence: hasSpace(building a,space b); | property: getProperty(building a,property b) | |
| **Full prompt** | All 8 categories | property: getProperty(element a,property b) ✓ | existence: hasElement(space a,element b) ✓ | property: getProperty(space a,property b) ✗ | geometry: getElementWidth(element a, type b) ✗ | |
| **Refined prompt** | property; existence; space_location | property: getProperty(element a,property b) ✓ | existence: hasElement(space a,element b); ✓ | existence: hasSpace(building a,space b); ✓ | property: getProperty(building a,property b); ✓ | property: getProperty(space a,property b) |

Fig. 10 Exemplary outputs of the proposed LLM-FuncMapper

## 5.3 Hallucination rates for LLMs

The hallucination rates for different models are presented in Table 5. Notably, Deepseek-R1 combined with a rule-based adaptive prompt exhibited the lowest hallucination rate, at only 0.14%, while Claude 3.5-Sonnet with a rule-based adaptive prompt also showed a relatively low rate of 0.29%, highlighting their superior reliability and minimal tendency for hallucination. In contrast, Deepseek-V3 showed the highest rate of hallucinations.

Table 5 Rate of hallucinations of different models

| Model | Method | Number of FF | Number of RF | Rate of hallucinations |
|---|---|---|---|---|
| ChatGPT4o | Full prompt | 18 | 1285 | 1.40% |
| | Rule-based adaptive prompt | 4 | 1273 | 0.31% |
| Claude 3.5-Sonnet | Full prompt | 6 | 1383 | 0.43% |
| | Rule-based adaptive prompt | 4 | 1402 | 0.29% |
| Deepseek-V3 | Full prompt | 28 | 1443 | 1.94% |
| | Rule-based adaptive prompt | 20 | 1407 | 1.42% |
| Deepseek-R1 | Full prompt | 4 | 1424 | 0.28% |
| | Rule-based adaptive prompt | 2 | 1380 | 0.14% |

### 5.4 F1 scores of LLMs in interpreting clauses of varying difficulty

Finally, the F1 scores of LLMs in interpreting clauses of varying difficulty are shown in Figure 11. Since the rule-based adaptive prompt engineering improves LLM performance, only the results of LLMs using this method are compared.

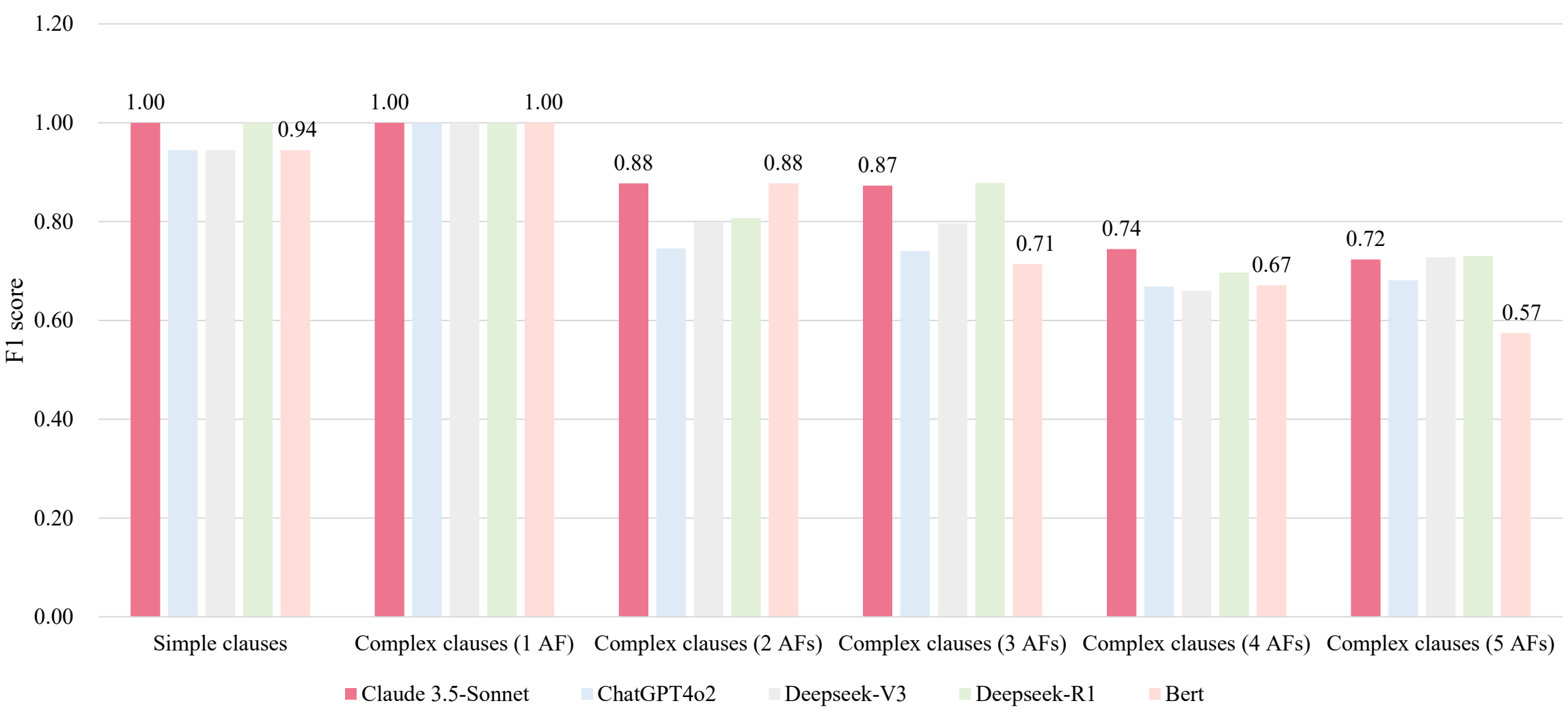

Fig. 11 Accuracy of LLMs in interpreting regulations of varying difficulty which AFs denote atomic functions; thus, "complex clauses (2 AFs)" refers to complex clauses composed of two atomic functions.

The results demonstrate that the proposed method achieves nearly perfect F1 scores (approaching 100%) on simple clauses and complex clauses with a single atomic function. For more complex clauses involving 2–3 atomic functions, the F1 score consistently exceeds 0.8, significantly outperforming fine-tuned BERT. However, the prediction accuracy decreases for highly complex clauses containing 4–5 atomic functions, highlighting the need for further research and improvements.

Through our testing, we discovered that Claude 3.5-Sonnet outperformed the others and demonstrated the ability to identify parts of the atomic functions. Consequently, Claude 3.5-Sonnet was selected as the model for further analysis. It should be noted that the function recommendation task is a very challenging end-to-end task. Because the parameter is an important component of the functions that specifies what kind of object will be input. Thus, the matched functions are only considered correct if the parameters are also predicted accurately. Such an evaluation method is more user-friendly for domain experts who may not be familiar with the atomic function library, but it also poses more significant prediction challenges for LLMs. Additionally, Claude 3.5-Sonnet is currently employed, and its training data does not encompass the proposed atomic function library and thus has certain limitations. Although the performance of the method is not so high, it already exceeds that of similar tasks such as API recommendation (Patil et al., 2023), and can be used to recommend proper predefined functions for domain experts and help them interpret regulatory clauses more quickly.

## 6 Proof of concept in ARC

In this section, the real application of the proposed method is demonstrated by performing rule checking on an actual plant building. The plant building has a total floor area of 6920 $m^2$ with two aboveground floors. The process begins by exporting an IFC model of the plant building from the Revit BIM model.

Next, two example clauses containing complex statements are selected from the *Chinese Code for fire protection design of buildings* (GB 50016-2014). These clauses involve counting the number of elements and calculating distances, respectively, as depicted in Fig. 12. The complex statements within these clauses are challenging to describe using conventional logical languages like first order logic.

To interpret these clauses, some of the defined atomic functions are implemented based on Python and IFCOpenShell (IFCOpenShell, 2023). Utilizing the proposed LLM-FuncMapper, suitable atomic functions are matched and utilized to interpret the selected clauses into Python code through the composition of atomic functions. LLM-FuncMapper produced function recommendations for a single clause in ~5 seconds, whereas domain experts typically required several minutes to analyze the same clause and select functions manually. While this is hardware- and clause-complexity-dependent, it indicates meaningful speedups in the recommendation step.

The "Mapped functions" parts in Fig. 12 (a)&(b) show the matched 3 functions for clauses 1 and 4 functions for clause 2, respectively. The functions marked in red are the correctly matched ones and are then used to interpret the clause into computer-executable Python code. As for clause 1, a total of three atomic functions are required, and two functions among the three matched functions are correct. As for clause 2, a total of four functions are required, and three functions among the four matched functions are correct. With the LLM-FuncMapper, domain experts spend less time searching for the proper functions. We then input the clauses to be interpreted and the required atomic functions into the LLM using a prompt, in order to automatically generate the model checking code. The code generated using Deepseek R1 and the prompts used are shown in Fig. 12. Then, the Python code can be executed to check the model. The checking results of the IFC model using the example clauses are also presented in Fig. 12.

As for clause 1, the fire protection zone with fewer than 2 safety exits that do not meet the requirement is identified. The global ID of the space associated with this violation is "1wXU_jTED61P7ZYfArwKmy". The model checker provides the global ID of the selected elements, enabling users to locate and address the identified elements accordingly.

As for clause 2, the clause specifies that the horizontal distance between the nearest two adjacent safety exits in each fire compartment should not be less than 5m. In the checked IFC model, the horizontal distance between two adjacent safety exits is measured to be 86.3m, satisfying the requirements of the clause.

This test validates the feasibility of executing atomic functions for automated rule checking. When engineers need to verify a model's compliance with a particular regulation, our method automatically generates the checking logic and corresponding code for all relevant clauses. The engineer simply selects the correct function from a few options generated by the LLM-FuncMapper, replaces the subject to be checked, and the system can then be used for fully automated model verification. This adoption of atomic functions and rule-based adaptive prompt engineering of LLM offers engineers the advantage of reduced difficulty in interpreting building codes, facilitating more efficient rule checking processes.

**Clause 1:**
The number of safety exits in the plant is not less than 2 for each fire protection zone.

**Matched functions:**
`getProperty(space a,property b); hasElement(space a, element b); getNumberOfElement(element a)`

**The prompt for code generation:**
Please help me convert the standard text into Python code for rule checking, based on ifcopenshell to extract properties. The functions involved in the checking process include getProperty(space a, property b) to retrieve property b of space a, and hasElement(space a, element b) to check if space a contains element b.
Standard text: The number of safety exits in the plant is not less than 2 for each fire protection zone.

**Generated executable codes:**
```
def getNumberOfElement_SafetyExit(ifc_space) -> int:
    doors = hasElement(ifc_space, 'IfcDoor')
    safety_exit_num = 0
    for door in doors:
        if getProperty(door, 'fire_exist?'):
            safety_exit_num += 1
    return safety_exit_num

ifc_spaces = ifcopenshell.open(ifc_file_path).by_type('IfcSpace')
fire_prevent_spaces = [space for space in ifc_spaces if getProperty(space, 'fire_protection_zone?')]
Pass = True
for space in fire_prevent_spaces:
     if getNumberOfElement_SafetyExit(space) <2:
         print(f'The space with id: {space.GlobalId} not pass the checking!')
         pass = False
if pass: print("Pass the check!")
```

**Result:**
The space with id: **1wXU_jTED61P7ZYfArwKmy** not pass the checking!

(a) clause 1

**Clause 2:**
The horizontal distance between the nearest two safety exits in each fire protection zone shall not be less than 5m.

**Matched functions:**
```
getElementDistance(element a,element b,type c); getProperty(element a, property b);
isAdjacent(space a, element b, direction c); hasElement(space a,element b)
```

**The prompt for code generation:**
Please help me convert the standard text into Python code for rule checking, based on ifcopenshell to extract properties. The functions involved in the checking process include getElementDistance(element a, element b, type c) to measure the distance between component a and component b using method c, getProperty(element a, property b) to retrieve the property b of component a, and hasElement(space a, element b) to check if space a contains component b.
Standard text: The horizontal distance between the nearest two safety exits in each fire protection zone shall not be less than 5m.

**Executable codes:**
```
def getMinSafetyExitDistance(ifc_space) -> float:
    doors = hasElement(ifc_space, 'IfcDoor')
    safety_exits = []
    for door in doors:
        if getProperty(door, 'fire_exist?'):
            safety_exits.append(door)
    safety_exits = list(set(safety_exits))
    min_distance = np.Inf
    for i in range(len(safety_exits)):
        for j in range(i + 1, len(safety_exits)):
            safety_exit_a = safety_exits[i]
            safety_exit_b = safety_exits[j]
            new_distance = getElementDistance(safety_exit_a, safety_exit_b)
            if new_distance < min_distance:
                min_distance = new_distance
    return min_distance

ifc_spaces = ifcopenshell.open(ifc_file_path).by_type('IfcSpace')
fire_prevent_spaces = [space for space in ifc_spaces if getProperty(space, 'fire_protection_zone?')]
Pass = True
for space in fire_prevent_spaces:
    if getMinSafetyExitDistance(space) < 5000:
        print(f'The space with id: {space.GlobalId} not pass the checking!')
        pass = False
if pass: print("Pass the check!")
```

**Result:**
Pass the check!

(b) clause 2

Fig. 12 Proof-of-concept of LLM-FuncMapper for automated compliance checking

# 7 Discussion

As for rule interpretation, some building codes encompass a large number of clauses with intricate semantics that rely on complex computational logics, including implicit properties inference and geometric calculations. However, the use of conventional logical representations like first-order logic proves insufficient in effectively interpreting these complex clauses into executable code due to a lack of domain knowledge and limited expressibility. In the context of ARC, accurate interpretation and transformation of such clauses into executable rule code is a critical step, as it directly determines the effectiveness, adaptability, and scalability of compliance checking systems. To enable the automated rule interpretation over a broader range of clauses, LLM-FuncMapper, an approach to match various regulatory clauses to predefined functions based on the LLM, is proposed. LLM-FuncMapper consists of atomic function library development and rule-based adaptive prompt engineering. To realize this, this work endeavors to establish a series of atomic functions to capture shared computational logics of implicit properties and complex constraints, through a systematic and meticulous analysis of the clauses in the *Chinese Code for fire protection design of buildings* (GB 50016-2014). In addition, we design a prompt template and a rule-based adaptive prompt tuning strategy for atomic function matching via LLMs. The established atomic functions and the corresponding atomic function matching method equip domain experts with practical techniques to enhance their rule interpretation process. Finally, the results of the conducted statistical analyses, experiments, and the proof of concept highlight several insights as follows:

Through a systematic and meticulous analysis of Chinese building codes, we establish an atomic-function library of 66 functions. To support multiple data models and languages, functions are specified at the interface level, since internal calculations differ across models. These functions capture shared computational logic in clauses and act as reusable building blocks for interpretation—the first such library for Chinese regulations. Usage shows a long-tail pattern, suggesting engineers should prioritize the development and implementation of high-frequency functions. Besides, the statistical analyses also reveal the great potential of the function to represent and interpret rules. Without high-order functions, 92.23% of computer-processable clauses cannot be fully interpreted; combining low- and high-order functions enables representation of nearly 100% of such clauses as executable code (Fig. 13). This indicates that although the low-order functions exhibit the highest frequency of usage, solely using the low-order functions can not sufficiently meet the requirements of fully automated rule interpretation.

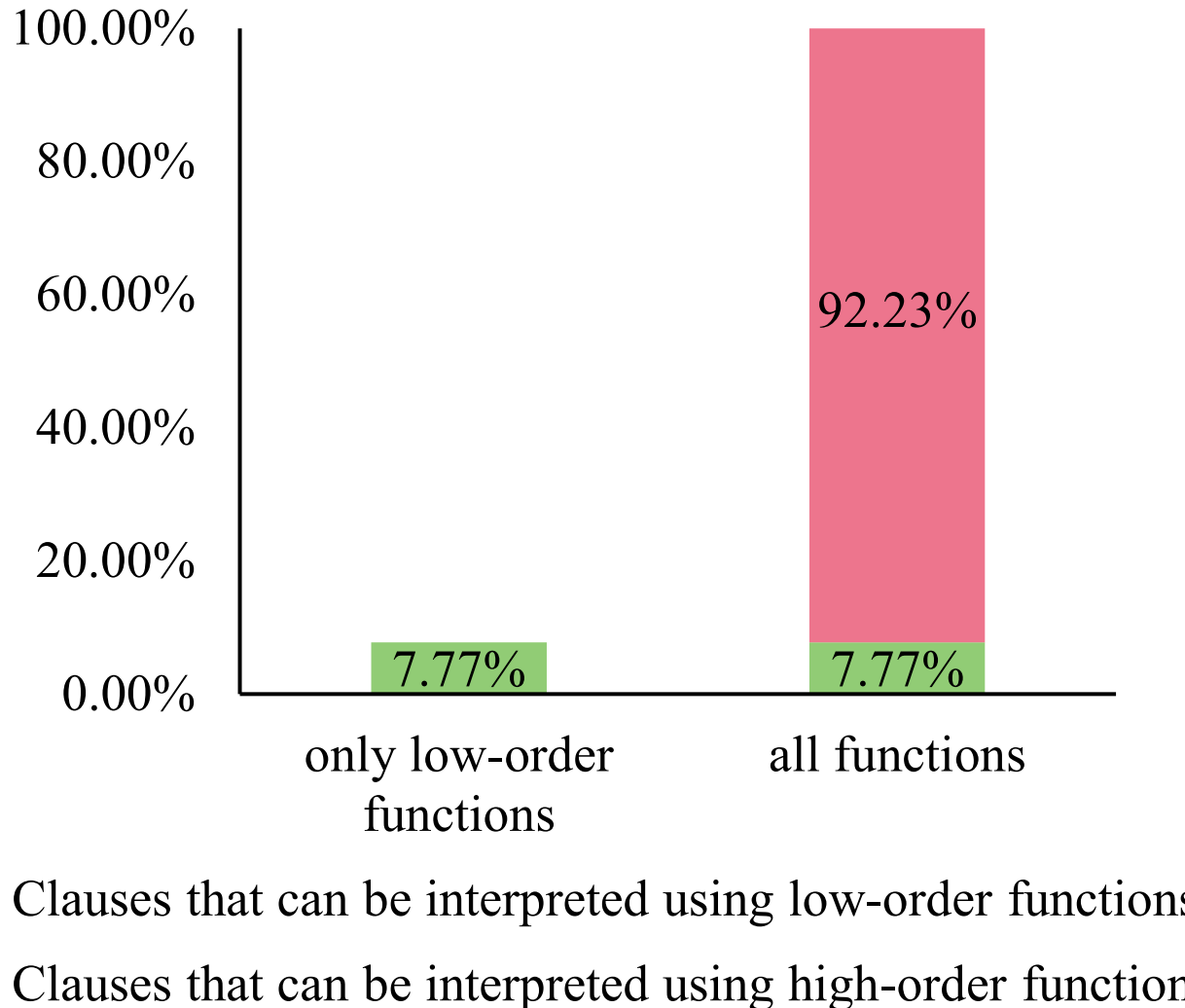

Fig. 13 Percentage of clauses that can be interpreted among all computer-processable clauses

The conducted experiments show that the designed prompt templates enable the common LLM for effective function matching. Then, with the proposed rule-based adaptive prompt tuning strategy, the rule-based adaptive prompt can further enhance the matching capabilities of LLMs, which improves the prediction recall rate by 4.6%, 5.6%, 1.2% and 2.8% for ChatGPT4o, Claude 3.5-Sonnet, Deepseek-V3, and Deepseek-R1, respectively. With this prompt, Claude 3.5 Sonnet outperforms a fine-tuned BERT by 19%, achieving substantial gains without finetuning. Besides, in terms of hallucination rates, Deepseek-R1 has the lowest hallucination rate, followed by Claude 3.5-Sonnet, ChatGPT4o, and finally Deepseek-V3.

Our proof of concept shows that the proposed methods ease the interpretation of complex clauses and that LLMs can attain a broad understanding of regulatory text. To our knowledge, this is the first attempt to introduce LLMs for interpreting complex design clauses into executable code, which may shed light on further adoption of LLMs in the construction domain. Potential applications include using LLMs to extract design information for intelligent/parametric design and to mine domain knowledge for specialized question answering.

The currently developed sentence-level LLM-FuncMapper can interactively identify functions for domain experts, effectively reducing the challenges in function selection. However, future advancements can be made by developing a word-level LLM-FuncMapper. By integrating the word-level LLM-FuncMapper into the existing automated rule interpretation framework, phrases containing computational logics can be matched to predefined atomic functions, enabling fully automated interpretation and code generation of a broader range of clauses.

Several aspects of the limitations of this research are identified and could be studied in the future:

The atomic function library can be expanded to encompass a broader range of domains. In this study, the atomic function library was constructed based on the Chinese Code for Fire Protection Design of Buildings (GB 50016-2014), which is sufficient to cover almost all interpretation needs related to fire protection clauses. Admittedly, when applying the proposed method to new domains, domain experts may still need to refine or extend the atomic function library. Nevertheless, many commonly used atomic functions, such as hasSpace(building a, space b) and getFloorArea(space a, type b), are highly reusable and appear across various regulations beyond fire protection. Therefore, experts only need to supplement a limited number of domain-specific functions (e.g., getFireProofDistance) when adapting the system to new codes. Once the function library of a particular domain is constructed, it can be directly reused across all regulations in that domain, ensuring scalability and efficiency. Future research can focus on enriching the atomic function library to accommodate the regulatory requirements of different domains. LLMs can assist this expansion: we can provide the existing catalog and the target regulation text to the LLM to help propose and extract the additional specialized atomic functions needed.

This study only examined the four most advanced LLMs available at the time. However, with ongoing advancements, these models could be replaced with more advanced LLMs in the future to achieve higher precision in function matching. Furthermore, as token costs decrease, the method proposed in this study will become more cost-effective and applicable to a wider range of scenarios.

For LLM-FuncMapper, scalability is a challenge: as the function catalog grows, prompt-based knowledge injection hits LLM token limits. General-purpose LLMs also underperform in specialized AEC contexts—during testing, we observed poor matches, hallucinated functions, and repetition of early-listed functions. Future work should explore fine-tuning domain-specific LLMs for AEC (Zheng et al., 2022b) and advancing prompt engineering/tuning to improve function matching.

In the current implementation, the atomic functions selected by the LLM are returned as a set of recommended candidates rather than in the precise sequence of application. This means that while the LLM can identify which functions are relevant, the sequencing of function calls requires either rule-based reasoning or further expert intervention. Future research could extend the framework to guide LLMs in inferring not only the relevant functions but also their sequential order, for example, by integrating chain-of-thought prompting or dependency-graph-based composition strategies.

In the current study, we deliberately filter qualitative, non-computer-processable clauses to keep the evaluation focused on verifiable checks; however, we agree that LLMs are promising for interpreting and quantifying such rules. As future work, we envision (i) rewriting vague terms into measurable proxies, for example, mapping “near” to a distance bound conditioned on occupancy, fire load, or egress context; and (ii) fuzzy induction that proposes candidate numerical ranges by retrieving analogs from related standards and historical approvals, with expert-in-the-loop confirmation. This pipeline would preserve traceability (original wording to threshold rationale) while enabling automated checking of clauses that are currently qualitative.

At present, our LLM component is limited to recommending atomic functions. In future work, we will automate code generation directly after function recommendation to realize a fully end-to-end pipeline.

The current rule-based inference method, which relies on keyword matching, can be further improved by adopting deep learning-based techniques to enhance its performance (e.g., higher precision and covering more categories). This can further reduce the number of atomic functions input into LLMs, thus shortening the prompt length and enhancing the matching performance.

# 8 Conclusion

As a vital stage of automated rule checking (ARC), converting regulatory texts into executable code requires significant effort. However, interpreting regulatory clauses with implicit properties or complex computational logic into code remains challenging due to the lack of domain knowledge and the limited expressibility of conventional logic representations. To address these challenges, through a systematic analysis of building codes, 66 atomic functions are defined to capture the shared computational logic of implicit properties and complex constraints. These atomic functions serve as fundamental blocks for rule interpretation and code generation. Then, LLM-FuncMapper is introduced as a large language model (LLM)-based approach that matches various regulatory clauses to predefined atomic functions. A prompt template incorporating a chain of thought (CoT) is developed and enhanced with rule-based adaptive strategies to optimize function matching performance. Finally, executable code is generated by composing functions through the LLMs. Statistical analyses reveal a long-tail distribution pattern of the developed functions. Certain functions emerge as particularly prevalent, prompting algorithm engineers to prioritize their development and implementation. Experiments demonstrate that LLM-FuncMapper, employing few-shot methods, achieves a performance of 81.55%, surpassing the fine-tuned BERT model's 68.50% by 19%, significantly reducing the reliance on extensive manually annotated data for model fine-tuning. Finally, the proof of concept in automated rule interpretation also demonstrates the possibility of LLM-FuncMapper in interpreting complex regulatory clauses into executable code through the composition of atomic functions.

This study addressed the challenge of interpreting regulatory clauses with implicit properties or complex computational logic and automatically generated checking logic and code for all relevant clauses, reducing the difficulty of interpreting building codes and facilitating more efficient rule-checking processes. Although GB 50016-2014 is used as an example, the method can be adapted to other regulations. Additionally, it can also be integrated into software as an API, translating engineers' requirements into executable functions for human-computer interaction. For instance, personalized requirements from engineers for model checks or design requirements from designers can be translated into executable functions and code, facilitating interactive building modification and design generation. Furthermore, to the best of our knowledge, this study is the first attempt to introduce LLMs for interpreting complex design clauses into executable code, which may shed light on the further adoption of LLMs in the construction domain. For instance, LLMs can be used to extract design information for supporting intelligent or parametric design, as well as to extract knowledge from massive documents of the construction domain to achieve domain-specific intelligent question-answering.

# APPENDIX

Table A presents a comprehensive overview of the atomic function library established in this study,

encompassing 66 atomic functions. The functions are classified into eight categories based on their usage, including existence, quantity, geometry, distance, area, space location, window-to-wall ratio, and property. Table A introduces the category, type of the return values, function name, and the corresponding description of each atomic function. The atomic function library consists of 5 low-order functions and 61 high-order functions, with the low-order functions distinguished by being highlighted in green.

Table A Overview of the atomic functions library

| CATEGORY | OBJECT | OUTPUT | FUNCTION_NAME | DESCRIPTION |
|---|---|---|---|---|
| Existence | Sapce | Boolean/ Collection | hasSpace(space a,space b) | Space a has space b |
| | Space& Element | Boolean/ Collection | hasElement(space a,element b) | Space a has element b |
| | Space& Equipment | Boolean/ Collection | hasEquipment(space a,equipment b) | Space a has equipment b |
| | Space& Building | Boolean/ Collection | hasBuilding(space a,building b) | Space a has building b |
| | Space& Goods | Boolean/ Collection | hasGoods(space a,goods b) | Space a has goods b |
| | Space | Collection | hasExternal(space a) | There are several external elements of space a |
| | Element | Boolean | hasElement(element a,element b) | Element a has element b |
| | Element& Equipment | Boolean | hasEquipment(element a,equipment b) | Element a has equipment b |
| | Equipment & Element | Boolean | hasElement(equipment a,element b) | Equipment a has element b |
| | Equipment & Goods | Boolean | hasGoods(equipment a,goods b) | Equipment a has goods b |
| | Building& Sapce | Boolean | hasSpace(building a,space b) | Building a has space b |
| | Building& Element | Boolean | hasElement(building a,element b) | Building a has element b |
| | Building& Equipment | Boolean | hasEquipment(building a,equipment b) | Building a has equipment b |
| | Building | Boolean | hasBuilding(building a,building b) | Building a has building b |
| | Building& Goods | Boolean | hasGoods(building a,goods b) | Building a has goods b |
| | Building | Collection | hasExternal(building a) | There are several external elements of building a |
| Quantity | Space | Integer | getNumberOfSpace(space a) | There are a number of space a |

| | Element | Integer | getNumberOfElement(element a) | There are a number of element a |
|---|---|---|---|---|
| | Equipment | Integer | getNumberOfEquipment(equipment a) | There are a number of equipment a |
| | Building | Integer | getNumberOfBuilding(building a) | There are a number of building a |
| Geometry | Space | Float | getSpaceWidth(space a, type b) | Space a has a width,which is measured by measurement type b; inner wall (type 0), center (type 1), outer wall (type 2) |
| | Space | Float | getSpaceHeight(space a, type b) | Space a has a height, which is measured by measurement type b; floor to ceiling (type 0), center to center (type 1), floor to floor (type 2) |
| | Space | Float | getSpaceLength(space a, type b) | Space a has a length, which is measured by measurement type b; inner wall (type 0), center (type 1), outer wall (type 2) |
| | Element | Float | getElementWidth(element a, type b) | Element a has a width, which is measured by measurement type b; inside (type 0), outside (type 1) |
| | Element | Float | getElementHeight(element a, type b) | Element a has a height, which is measured by measurement type b; inside (type 0), outside (type 1) |
| | Element | Float | getElementLength(element a, type b) | Element a has a length, which is measured by measurement type b; inside |
| | | | | (type 0), outside (type 1) |
| | Element | Boolean | isOpened(element a) | Does element a have several openings |
| | Element | Boolean | isEvenOpened(element a) | Is element a even opened |
| | Equipment | Float | getEquipmentWidth(equipment a, type b) | Equipment a has a width, which is measured by measurement type b; inside (type 0), outside (type 1) |
| | Equipment | Float | getEquipmentHeight(equipment a, type b) | Equipment a has a height, which is measured by measurement type b; inside (type 0), outside (type 1) |
| | Equipment | Float | getEquipmentLength(equipment a, type b) | Equipment a has a length, which is measured by measurement type b; inside (type 0), outside (type 1) |
| | Building | Float | getBuildingWidth(building a, type b) | Building a has a width, which is measured by measurement type b; inside (type 0), outside (type 1) |
| | Building | Float | getBuildingHeight(building a, type b) | Building a has a height, which is measured by measurement type b; inside (type 0), outside (type 1) |
| | Building | Float | getBuildingLength(building a, type b) | Building a has a length, which is measured by measurement type b; inside (type 0), outside (type 1) |
| Distance | Sapce | Float | getSpaceDistance(space a, space b, type c) | The distance between space a and space b is measured by |

| | | | |
|---|---|---|---|
| | | | the measurement type c; door to door (type 0), shortest (type 1) |
| Space& Element | Float | getLinearDistance(space a, element b) | The distance between a random spot in space a and element b is measured by the measurement criteria type c; shortest (type 0), longest (type 1) |
| Element | Float | getElementDistance(element a, element b, type c) | The distance between element a and element b is measured by the measurement criteria type c; outside (type 0), centerline (type 1) |
| Element& Building | Float | getDistance(element a, building b, type c) | The distance between element a and building b is measured by the measurement criteria type c; outside (type 0), centerline (type 1) |
| Element& Equipment | Float | getDistance(element a, equipment b, type c) | The distance between element a and equipment b is measured by the measurement criteria type c; outside (type 0), centerline (type 1) |
| Equipment | Float | getEquipmentDistance(equipment a, equipment b, type c) | The distance between equipment a and equipment b is measured |

| | | | | |
|---|---|---|---|---|
| | | | | by the measurement criteria type c; outside (type 0), centerline (type 1) |
| | Building | Float | getBuildingDistance(building a, building b, type c) | The distance between building a and building b is measured by the measurement criteria type c; shortest (type 0), longest (type 1), door to door (type 2) |
| | Building | Float | getFireproofDistance(building a, building b) | The fireproof distance between building a and building b |
| Area | Space | Float | getFloorArea(space a, type b) | Area of space a is measured by the measurement criteria type b; inside (type 0), center line (type 1), outside wall (type 2) |
| | Element | Float | getFloorArea(element a, type b) | Area of element a is measured by the measurement criteria type b; inside (type 0); outside (type 1) |
| | Building | Float | getFloorArea(building a, type b) | Area of building a is measured by the measurement criteria type b; inside (type 0), center line (type 1), outside wall (type 2) |
| Space location | Space | Boolean | isAccessible(space a, space b, type c, direction d) | Space a is able to move to space b through |

| | | | |
|---|---|---|---|
| | | | moving route type c and direction d; door to door (type 0); others (type 1) |
| Space | Boolean | isVisibleFrom(space a, space b, type c) | Space b is visible from space a through observation route type c; open door (type 0), window (type 1) |
| Space | Boolean | isAdjacent(space a, space b, direction c) | Space a is adjacent space b in direction c; |
| Space& Element | Boolean | isAdjacent(space a, element b, direction c) | Space a is adjacent element b in direction c; |
| Space& Equipment | Boolean | isAdjacent(space a, equipment b, direction c) | Space a is adjacent equipment b in direction c; |
| Element | Boolean | isOpenDirection(element a, type b) | The opening direction of component a is type b; inner (type 0), outer (type 1) |
| Element | Boolean | isFacedDirectly(element a, element b) | Element a is faced directly to element b |
| Space | Collection | getAccessibleWithSpace(space a, direction b, type c) | Get the space that is accessible with space a through moving route type c and direction b; door to door (type 0); others (type 1) |
| Equipment | Boolean | isConnectedTo(equipment a, equipment b) | Equipment a is connected to equipment b |
| Equipment | Boolean | isGroupArranged(equipment a, equipment b) | Equipment a and equipment b are group arranged |
| Building | Boolean | isAccessible(building a, building b, type c, direction d) | Building a is able to move to building b through |

| | | | | |
|---|---|---|---|---|
| | | | | moving route type c and direction d; via corridor (type 0), door to door (type 1) |
| | Building | Boolean | isAdjacent(building a, building b, direction c) | Building a is adjacent building b in direction c; |
| | Building& Space | Boolean | isAdjacent(building a, space b, direction c) | Building a is adjacent space b in direction c; |
| | Building& Equipment | Boolean | isAdjacent(building a, equipment b, direction c) | Building a is adjacent equipment b in direction c; |
| | Building | Boolean | isGroupArranged(building a, building b) | Building a and building b are group arranged |
| Window wall ratio | Element | Float | getWindowWallRatio(element a, element b, orientation c) | This is the ratio of element a and element b in the four cardinal orientation |
| Property | Space | String/Float / Boolean/ Collenction | getProperty(space a, string b) | Gets the property b of space a; |
| | Element | String/Float / Boolean/ Collenction | getProperty(element a, string b) | Gets the property b of element a; |
| | Equipment | String/Float / Boolean/ Collenction | getProperty(equipment a, string b) | Gets the property b of equipment a; |
| | Building | String/Float / Boolean/ Collenction | getProperty(building a, string b) | Gets the property b of building a; |
| | Goods | String/Float / Boolean/ Collenction | getProperty(goods a, string b) | Gets the property b of goods a; |

Table B presents the full set of predetermined keywords for rule categories.

Table B Full set of keywords for rule categories

| Category | Keywords (in Chinese) |
|---|---|
| quantity | ‘个’, ‘扣’, ‘次’, ‘数量’, ‘部’, ‘排数’, ‘座’, ‘樘’ |
| geometry | ‘宽度’, ‘长度’, ‘高度’, ‘较高’, ‘较低’ |
| distance | ‘距’, ‘距离’, ‘疏散距离’, ‘间距’ |
| area | ‘面积’, ‘体积’ |